 \documentclass[final,5p,times]{elsarticle}
\usepackage{graphicx}
\usepackage{amssymb}
\usepackage{caption}
\usepackage{subcaption}
\usepackage{rotating}
\usepackage{setspace}

\usepackage[switch]{lineno}

\usepackage[colorlinks=true]{hyperref}
\hypersetup{colorlinks=true,linkcolor = blue,urlcolor  = blue,
            citecolor = blue,
            anchorcolor = blue}
            
\usepackage{array}
\newcolumntype{L}{>{\centering\arraybackslash}m{3cm}}

\journal{ISPRS Journal of Photogrammetry and Remote Sensing}

\begin{document}

\begin{frontmatter}
\title{Crop identification using deep learning on LUCAS crop cover photos}

% Author Orchid ID: enter ID or remove command
%\newcommand{\orcidauthorA}{0000-0002-7326-7684} % Add \orcidA{} behind the author's name
%\newcommand{\orcidauthorB}{0000-0002-2734-4538} % Add \orcidB{} behind the author's name
%\newcommand{\orcidauthorC}{0000-0002-9103-7081} % Add \orcidC{} behind the author's name

% RD http://orcid.org/0000-0002-7326-7684
% GL http://orcid.org/0000-0002-2734-4538
% MV https://orcid.org/0000-0002-9103-7081

\author{Momchil Yordanov  $^{1}$, Rapha\"{e}l d'Andrimont $^{1}$, Laura Martinez-Sanchez $^{1}$, Guido Lemoine $^{1}$,\\
Dominique Fasbender $^{1,2}$,  Marijn van der Velde $^{1}$}
\address{ $^{1}$\quad European Commission, Joint Research Centre (JRC) {---} Food Security Unit, Ispra , Italy \\
 $^{2}$\quad Walloon Institute of Evaluation, Foresight and Statistics (IWEPS), Belgium 
}

\begin{abstract}
Crop classification via deep learning on ground imagery can deliver timely and accurate crop-specific information to various stakeholders. Dedicated ground-based image acquisition exercises can help to collect data in data scarce regions, improve control on timing of collection, or when study areas are to small to monitor via satellite. Automatic labelling is essential when collecting large volumes of data. One such data collection is the EU's Land Use Cover Area frame Survey (LUCAS), and in particular, the recently published LUCAS Cover photos database. The aim of this paper is to select and publish a subset of LUCAS Cover photos for 12 mature major crops across the EU, to deploy, benchmark, and identify the best configuration of Mobile-net for the classification task, to showcase the possibility of using entropy-based metrics for post-processing of results, and finally to show the applications and limitations of the model in a practical and policy relevant context. In particular, the usefulness of automatically identifying crops on geo-tagged photos is illustrated in the context of the EU's Common Agricultural Policy. The methodology makes use of crop calendars from various sources to ascertain the mature stage of the crop; of the PyGeon library for labelling of image quality; of an extensive paradigm for hyper-parameterization of Mobile-net from random parameter initialization; and of various techniques from information theory in order to carry out more accurate post-processing filtering on results. The work has produced a dataset of 169,460 images of mature crops for the 12 classes, out of which 15,876 were manually selected as representing a clean sample without any foreign objects or unfavorable conditions. The best performing model achieved a Macro F1 (M-F1) of 0.75 on an imbalanced test dataset of 8,642 photos. Using metrics from information theory, namely - the Equivalence Reference Probability, resulted in achieving an increase of 6\%. The most unfavorable conditions for taking such images, across all crop classes, were found to be too early or late in the season. The proposed methodology shows the possibility for using minimal auxiliary data, outside the images themselves, in order to achieve a M-F1 of 0.817 for labelling between 12 major European crops.
\end{abstract}


\end{frontmatter}

%\linenumbers
%\doublespacing
% \newpage
% \tableofcontents
% \newpage
\section{Introduction}
\label{S:1}
% - intorducing the general area of study
% - ground the paper within the area of study in terms of the flow of the introduction should follow the flow of the paper
% write many different research questions and funnel them down to the main ones
% image recognition -> env domain -> agri domain -> policy context
% context block for image collection and satellite images
% think of the figures

%The advent of Machine Learning (ML) techniques has fundamentally reshaped the way research is designed, conducted, and operationalized. It can hardly be stressed enough exactly how deep these methods have proliferated in realms as distant from each other as that of the scientific to the everyday. The affect of ML, and its offspring of Deep Learning (DL), on the scientific paradigm has been profound through concepts such as 'data deluge' (\citet{hey2003data}) and the 'end of theory' (\citet{anderson2008end}) and their respective technical realizations. Under this heading, and with increasing abundance of use cases, we see 

The Deep Learning (DL) paradigm is regarded as the Gold Standard of the Machine Learning (ML) community (\citet{alzubaidi2021review}). While there is understandably a trade-off between the better performance in the model and the amount of data and resources necessary, it is clear that there is a significant improvement with DL methods, especially so in the field of image classification. Recent advancements in Convolutional Neural Networks (CNNs) have made popular classification tasks ever more affordable in an operational context. 

Related to this is the ability to perform on-device processing in order to provide an option to anyone wanting to implement the technology, while keeping computational overhead low. A leading architecture in this regard is MobileNet (\citet{howard2017mobilenets}), of which there are third generation flavours (\citet{howard2019searching}), which in turn are both significantly smaller than and equal in performance to the previous ones (\citet{sandler2018mobilenetv2}). MobileNets are convenient, as they perform on par with other state of the art architectures such as Inception V3 on popular benchmarking datasets, but have up to 20 million less parameters. 

Another important point in making DL models operational is the proper and appropriate use of post-processing techniques. What is generally understood here are all manipulations done to the data as output from the model. In an image classification setting this means everything done on the data after the output of the softmax activation function, which is a probability vector with a value for each class, that sums to one (\citet{salvi2021impact}). Popular post-processing approaches include combining Random Forest or Support Vector Machine classifiers after the CNN output, majority voting (\citet{d2022monitoring}), patch aggregation (\citet{matvienko2020bayesian}), and thresholding. Thresholding is one of the most popular choices, as it is usually simple to implement - keep only the examples, for which the network has a Maximum Probability (MP) for the winning class higher than the threshold. An interesting development in the field is the re-mapping of the base probabilistic output to a value of higher versatility, taking notions from information theory such as Shannon information and entropy (\citet{bogaert2017information}).

%In the environmental domain, narrow artificial intelligence (AI) is being developed for e.g. better water resource management in the Baltic Sea by employing drones and machine vision (\citet{finlandCVdrone}), for tree species mapping (\citet{tanglarge}), and in the development of tools to better predict future changes in the climate (\citet{reneecho}). 

%With the world's population projected to increase to reach 11.2 billion by the end of the century (\citet{maxroserpopgrowth}), it is evident that there will be an equivalent rise in the need for food production. It thus becomes ever more important to find better and more efficient ways to produce food; a process which necessitates the monitoring of the entire production chain of food in order to safeguard its security. 

In the agricultural sector, these technological developments are reflected in projected increases of the use of Artificial Intelligence (AI) throughout the food production chain (\citet{louiscolumbus}). DL-aided Computer Vision (CV) in particular is crucial for automation and robotic tasks that rely on inspection, evaluation, and execution of management interventions (\citet{tian2020computer}). Ultimately, these technical innovations should contribute to decreasing costs, while increasing resource use efficiency and precision, of food production systems. An important element of the application of these technologies relates to the possibility of new ways of information exchange among the various actors in the food production chain. This may relate to certification of management practices (\citet{santoso2021machine}), traceability of products (\citet{kollia2021ai}), as well as communication towards consumers (\citet{zhu2021deep}), or indeed various activities in the realm of citizen science and food related topics (\citet{schiller2021deep}), including biodiversity (\citet{affouard2017pl}). In technical terms, the possibilities have already been successfully tested for weed management (\citet{wu2021review}), crop disease recognition and management (\citet{mohanty2016using}), and harvesting operations (\citet{kapach2012computer}). 

Activities also focus on training data collection and curation for increasingly specific applications. \citet{lu2020survey} identified 34 public image datasets collected under field conditions of relevance to precision agriculture. \citet{zheng2019cropdeep} presented a crop dataset for deep-learning-based classification and detection in precision agriculture, while \citet{sudars2020dataset} presented a dataset of annotated food crops and weed images for robotic CV control. In the Earth Observation (EO) domain, datasets such as CropHarvest (\citet{tseng2021cropharvest}) with more than 90,000 worldwide geographically diverse samples with labels, and the LUCAS 2018 Copernicus polygons (\citet{d2021lucas}) with almost 60,000 stratified samples in the EU, demonstrate the push from the community to have open and free data to facilitate ML-and-DL-driven research. In this manuscript, we focus on recognizing crops and rely on a selection of legacy close-up photos taken during five tri-annual Land Use/Cover Area frame Survey (LUCAS) surveys from 2006 to 2018 in the EU (\citet{d2022lucas}) for our training set. 

 A fitting application in this sense and specifically so in the European context is the ability of technology to deliver to the needs of regulating bodies that administer the technical regulations of the European Union (EU)'s Common Agricultural Policy (CAP). The CAP is the single largest item on the EU budget, amounting to a total of 58,38 billion euros in 2022, including funds allocated for rural development, market measures, and income support (\citet{caphomepage}). Thus developing technology for automatizing the application process and the evidence provision for practices required under subsidy schemes is in great demand by paying agencies of the respective Member States (MS).

While Copernicus Sentinel based monitoring of the CAP area subsidies is being developed and implemented (\citet{devosconceptual}),  ground based information in the form of geo-tagged pictures (\citet{sima2020use}) can support and complement the Checks by Monitoring approach (CbM). CbM relies on Copernicus Sentinel data-streams providing wall-to-wall coverage of EU territory and cloud-based processing on the Data and Information Access Services (DIAS) platforms. By using Copernicus Application Ready Data (CARD), in conjunction with geospatial information from the Land Parcel Identification Systems (LPIS) and Geo-Spatial Aid Applications (GSAA), it is possible to extract parcel-level information of markers (see \citet{devosconceptual}). These markers evidence specific practices (e.g. mowing, irrigation, etc.) that can be related to compliance requirements. Nevertheless, in situations that the Sentinel based checks do not lead to conclusive results, geo-tagged pictures can be used to support and complement checks. Such processing chains may have to be developed for each specific agri-environmental practice for which evidence is needed. In the current CAP programming period (2023-2027), this includes practices under GAEC (Good Agricultural and Environmental Conditions) conditionality, as well as eco-schemes and agri-environmental and climate measures.

\subsection{Objectives}
\label{sec:objectives}
The aim of the research is to benchmark and test computer vision models to recognize Major and Mature European Crops (MMEC) on close-up photos in a practical agricultural policy relevant context. Specific objectives are:
\begin{itemize}
\item To select and publish a subset of LUCAS cover photos representative for major and mature crops across the EU for training purposes. 
\item To deploy and benchmark a set of Mobile-net computer vision models to recognize crops on close-up pictures and identify the best performing model. 
\item To explore the use of probability and entropy-based metrics to threshold and filter correct and incorrect classifications.
\item To illustrate the applications and limitations of the model for inference in a practical and agricultural policy relevant context. 
\end{itemize}

\section{Materials and Methods}
\begin{figure*}[!h]
    \centering\includegraphics[width=\linewidth]{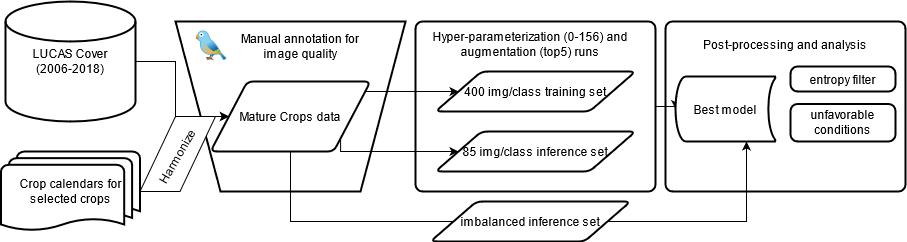}
\caption{Conceptual diagram of the study. The used data is shown on the left. LUCAS attributes are fused with harmonized crop calendars for the selected crops, after which the combined dataset undergoes a process of manual annotation using the pyGeon library. After annotating enough images of sufficiently high quality, a stratified sample across EU countries is done to select the training and inference sets, followed by the DL paradigm (described further in Figure \ref{fig:dl}). The DL workflow produces a best parameterized model, that in turn is used to inference over a large imbalanced set, where post-processing and further operational-context work takes place.}
\label{fig:fd}
\end{figure*}

%The study makes use of the LUCAS Cover database (REF), the crop calendars for the selected major European crops from various sources (Table \ref{tab:ccsource}), the output maps from Lorenzo's model (REF), and the use of expert knowledge to fill the gaps.
The methodological approach in the manuscript consists of 1) the procedure to select close-up LUCAS cover MMEC photos; 2) training, validating, and testing of a large set of Mobile-net based computer vision models; 3) applying the best model to inference photos across the EU; 4) evaluate model performance using metrics, derived from information theory to filter and understand why photos are not classified well; 5) test model performance against images exhibiting a series of unfavorable/out-of-scope conditions; 6) illustrate practical implications for protocol development. More specifically the workflow is presented in Figure \ref{fig:fd}.

% \begin{figure*}[!h]
%     \centering\includegraphics[width=\linewidth]{figures/LUCASVISION_flowDiagram_2.drawio.png}
% \caption{Flow diagram 2.}
% \label{fig:fd_2}
% \end{figure*}

%#methodology figures (hierarchically)
%- images of typical lucas crop pictures that consist in the set
%- map of the geographical distribution of the lucas images in train_ok with colored crops
%- flow diagram of the whole study
%   - LUCAS photos -> CC -> Bulletins -> Mature Crops -> ML (3 nets) (AWS) -> train and validation -> test inferecing (operational part) (suitable and non suitable pictures)

%To summarize what I understood in terms of terminology and what we’ll do for the paper (I had to hammer on again about the terminology/process to understand better!). 1) ‘Training and validation’ on the 400 images. During the optimization in hyperparameter space a ‘tiny’ split is made (e.g. 10\%) resulting in the validation accuracies (the validation) after ‘forward pass’ after final epoch for each model. 2) ‘Testing’, on the 100 images that still need to be selected - here we will calculate the ‘real’ accuracies, exposing the model to the images representative for each class. 3) The application in real-life. We will make a selection of e.g. 50 images for e.g. 4 different types of imagery (images representative as seen by model (cf step 2), with expected good performance; not suitable images, crop images out of ‘recognizable stage; images with objects on them (those you filtered out); images not properly taken (e.g. out of focus, too far away).

\subsection{Data}
\subsubsection{LUCAS cover photos}
%Paragraph on LUCAS and the LUCAS cover photos.

\begin{figure*}[!h]
    \centering\includegraphics[width=\linewidth]{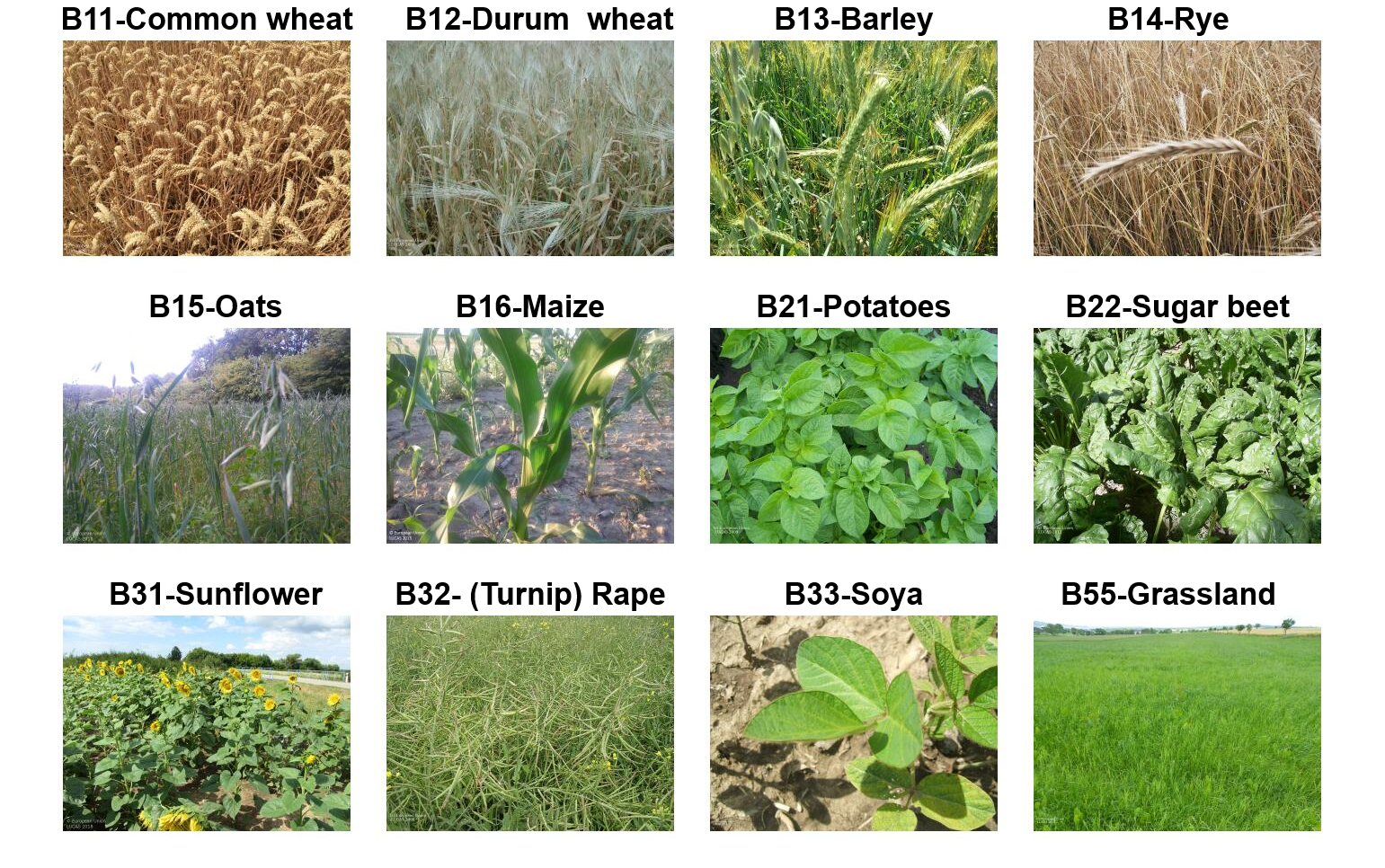}
\caption{Example of one image per class for the selected 12 major European crops.}
\label{fig:1perclass}
\end{figure*}

LUCAS Cover is a part of the core LUCAS survey since its inception and accordingly data has been collected for all five campaigns form 2006 to 2018. A total of 875,661 LUCAS Cover photos have been collected and 874,646 of those were published after anonymization  and curation \citep{d2022lucas}. In contrast to other LUCAS core imagery (four N,S,W,E, photos in the cardinal directions, and the point photo P), the Cover (C) photos, by protocol, must show the cover on the ground at the GPS location where the survey is carried out in such a way that the relevant crop, or plant can easily be identified during data quality controls. An example of one photo per selected crop is shown in Figure \ref{fig:1perclass}. The selection was done by reference to the main crops that are monitored and forecast by the European Commission's Joint Research Centre crop forecasting activities (AGRI4CAST, formerly MARS, see \citet{vanderVelde2019}). Omitting some classes due to data insufficiency, and including Temporary grassland, the number of crops arrives to 12. These are: Common wheat (B11), Durum wheat (B12), Barley (B13), Rye (B14), Oats (B15), Maize (B16), Potatoes (B21), Sugar beet (B22), Sunflower (B31), Rape and turnip rape (B32), Soya (B33), Temporary grassland (B55). The LUCAS Cover dataset is one of the two main inputs to the study, as shown on the left in Figure \ref{fig:fd}.

\subsubsection{Crop calendars and harmonization}
\label{sec:ccs}
One of the objectives of the study is the identification of mature crops on geo-tagged LUCAS imagery. The rationale being that, from an operational standpoint, the mature stage of the crop is the one in which it is most recognizable. The mature stages of the selected crops have to be firstly ascertained. One way of doing so is by collecting all crop calendars from the variety of sources available, harmonizing them into a common format, extracting the harvest period for each crop, and finally, through the use of expert knowledge, derive the pre-harvest mature stage of the crop.

A Crop Calendar (CC) is a schedule that provides timely information about crops in their respective agro-ecological zone. They are usually provided in tabular or gridded form and cover the space of a calendar year by dividing it into the planting, vegetative, and harvest stages of the respective crop. For the present purposes, CCs were gathered from various sources (Table \ref{tab:ccsource}) and harmonised  to a common style (AGRI4CAST), as it already hosts the data in tabular and numeric format, facilitating further processing.  It must be noted that certain steps had to be taken to account for instances where more than one variety (spring/winter, or early/late ware varieties) of the same crop is cultivated in a country. The decision was made to exclude countries that cultivate both varieties, and use the CC information for only those countries that cultivate the winter and early ware variety, with the information for the excluded countries being populated by expert knowledge.

% latex table generated in R 3.4.4 by xtable 1.8-4 package
% Mon Mar 29 15:10:00 2021
\begin{table}[ht]
\centering
\caption{Crop calendar sources and references}
\label{tab:ccsource}
\begin{tabular}{p{1.8cm}|p{3cm}|p{1.5cm}}
  \hline
\textbf{CC source} & \textbf{Crop} & \textbf{Link} \\ 
  \hline
 \citet{agri4cast} & Corn, Winter Wheat, Durum Wheat, Rice  & \href{https://agri4cast.jrc.ec.europa.eu/DataPortal/Index.aspx?o=}{AGRI4CAST}  \\ 
  \hline
  \citet{usgscc} & Sunflower, Barley, Rye, Soybeans, Spring Wheat, Rapeseed, Oats & \href{https://ipad.fas.usda.gov/rssiws/al/crop\_calendar/europe.aspx}{USDA}  \\ 
    \hline
  \citet{europabio} & Potato, Sugar Beet & \href{https://www.europabio.org/sites/default/files/120605\_report\_eu\_farming\_practices\_potato.pdf}{EUROPABIO potato}, \href{https://www.europabio.org/sites/default/files/120604\_report\_eu\_farming\_practices\_sugar\_beet.pdf}{EUROPABIO sugarbeet} \\ 
   \hline
\end{tabular}
\end{table}

\subsubsection{Expert knowledge gap filling and mature pre-harvest stages}
\label{sec:experts}
After harmonizing the CCs and extracting harvest stages at national level, the study fills the gaps and validates the result by means of expert knowledge. One way of identifying gaps is using the information from the JRC MARS bulletins (\citet{agri4castBullArchive}). These bulletins offer information on crop growth conditions and yield forecast at EU level and neighbouring countries like the UK, Ukraine, Black Sea area, and Maghreb. The rationale here being that if there is information in the bulletin about the yield of a certain crop for a certain country, then the crop is obviously cultivated in the respective country, and ergo - CC information about it should be present. After identifying the gaps, they were filled with all available information, comprising of interpolations from the COST 725 phenology network (\citet{koch2005cost}), and expert knowledge. A breakdown of all the information gathered and the sources it was collected from is available in the Supplementary Material Figure \ref{fig:finalTableHarvestStages}. The final step is acquiring the mature, pre-harvest conditions of the crops. This was accomplished again with the use of expert knowledge and was conducted in accordance with the following rules: for cereals, rapeseed, sunflower, and soya - remove the last half month and then add 2 months at the beginning of the harvest stage; for potatoes, sugarbeet, maize, and rice - remove the last half month and add 3 months.

\subsubsection{Manual photo pre-processing by visual assessment}
\label{sec:manual}
The dataset was then visually assessed with the use of the PyGeon jupyter library to remove examples not suited for the study. The photos were selected on the basis of what one could expect in farmer photos:  artificial background (map, hand, leg, pivot), and low quality photos (e.g. against the sun, shadowed, etc.) were not allowed. Close ups showing individual leaves, ears, grains were also removed. Overview photos where the crop appears somewhere in the background, usually mixed with other elements (a road, neighbouring field etc) were also removed. Photos with seeds only on a bare soil background (in cereals, soy, maize mostly), and other obviously wrong photos were also eliminated, although this happened only a few times. 

\begin{figure*}[!h]
    \centering\includegraphics[width=\linewidth]{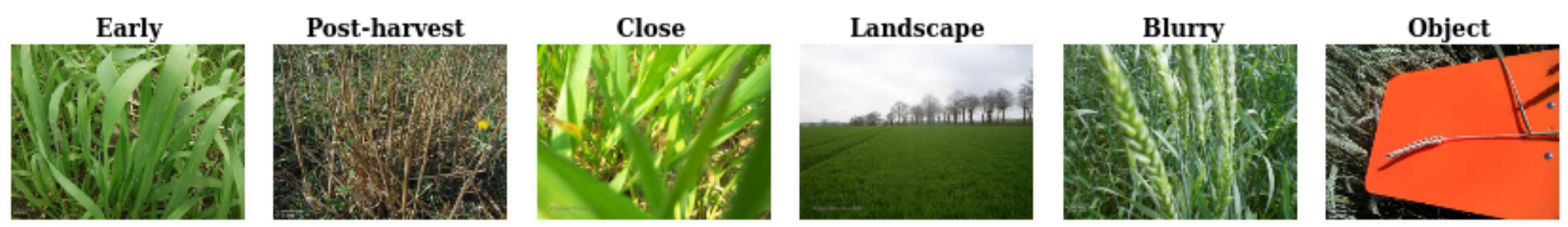}
\caption{The six classes of unfavorable conditions for Common wheat (B11).}
\label{fig:badimgs}
\end{figure*}

At this stage, photos flagged as not suitable to train on are later manually classified into one of the six categories of unfavorable conditions. These are out of season (too early or late in the season), out of protocol (too close or too far away to image the plant matter adequately), or either being blurred or there being a foreign object in the photo. A total of 354 images were selected in this way, while making sure that there is at least one photo per year, per LUCAS land cover class, per unfavorable condition. An example of unfavorable conditions for Common wheat (B11) is shown on Figure \ref{fig:badimgs}.

The final clean dataset used in this study is available for download \url{https://jeodpp.jrc.ec.europa.eu/ftp/jrc-opendata/DRLL/LUCASvision/}.

\subsection{Method}
\begin{figure*}[!h]
    \centering\includegraphics[width=\linewidth]{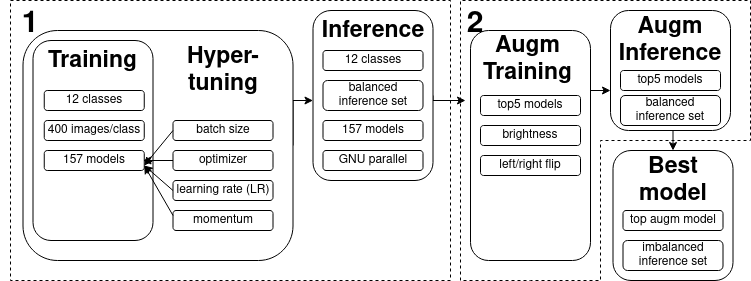}
\caption{Deep learning processing chain. The two rounds of training and inferencing are shown in black contour and labelled accordingly. }
\label{fig:dl}
\end{figure*}

The study makes use of a CNN for an image-classification CV exercise with a balanced training and inference set. There are two rounds of training and parallelized inferencing that make up the hyper-parameterization workflow (Figure \ref{fig:dl}) - one without and one with data augmentations (flip, brightness, etc).  After the final augmented inference, the best model is identified and it is fed with a much larger imbalanced inference set, supposed to represent a quasi-operational scenario. Specific and innovative post-processing techniques are also explored. 

\subsubsection{Training and inference set(s) sample selection} 
\label{sec:sets}
The selected number of photos per class for training was set to 400, following the current State of the Art \citep{soekhoe2016impact}. In order to select the set, a stratified sample across NUTS0 regions in the EU is made from the MMEC dataset (from Section \ref{sec:res_dataset}). This is done with the idea of having equal representation across EU countries, which allows for articulating conclusions on the European scale. 

In order to shorten processing time, instead of using the entire leftover (post-training-set-selection) set of images for inferencing, the study makes use of a custom inference set, sampled out of the leftover set. A total of 85 (the total number of examples of the least represented class (B12)) images per class were selected with a geographic distribution that matches the one of the training set. This ‘‘balanced" inference set was used during the first and second stage of inferencing (Figure \ref{fig:dl}).

The last set of images to be discussed is the ‘‘imbalanced" inference set, which includes all the photos left after the training set selection with all classes capped at 1000 examples per class. This set includes the previous ‘‘balanced" set and it is the one used on the identified best model in order to judge the possibility of using the model as an operational tool. It is also this set that any further developments are tested on.

\subsubsection{Hyper-parameter search and best model selection}
The network in use is MobileNet V2. The images vary in their native resolution (see Supplementary Material Table \ref{tab:resolultions}), but every image in the training and inference set is re-scaled to the net input size of 224x224. The effects of this re-scaling are discussed in section \ref{sec:recommendations}. The V2 MobileNets are trained for 3000 epochs, with the following settable parameters - learning rate, momentum, optimizer, batch size. These variables were experimented within a random space \citep{bergstra2012random} to generate values for initializing the learning process. In this way 157 model configurations were tried in order to find the best approximation for solving the problem. Model performance was then tested by carrying out an independent inference exercise on the dedicated balanced set. The models are then ranked based on their Overall Accuracy (OA) to find the top five performers. This completes the first round of training.

For the second round the top five performers are run through another cycle of training with the same configuration, but adding image augmentations, in this case random brightness and horizontal image flips. The same inferencing on the balanced set is done to rank the augmented models based on OA. The best performing of these is then taken as the overall best model.

\subsubsection{Operational use}
After the best model is identified, it is used on the imbalanced inference set (see Section \ref{sec:sets}). Because of the class imbalance, it was necessary to use a different metric - Macro-F1 (M-F1) \citep{opitz2019macro}. It is the results from this inference run that are presented in Section \ref{sec:results}. It is also on these results that the effectiveness of innovative post-processing techniques will be tested and upon which all the discussion will be carried out.

\subsubsection{Computational Infrastructure}
All the code developed for this study is available openly on the following repository: \url{https://github.com/Momut1/lucasVision}. The working environment was carried in a docker image. The processing pipeline is fully reproducible and automated to work by calling shell scripts that respectively carry out the hyper-tuning, inferencing, results derivation, and post-processing and plotting. For more information consult the \textit{readMe} of the git repository. The processing was done on the JRC BDAP, an in-house, cloud-based, versatile, petabyte-scale platform for heavy-duty processing \citep{soille2018versatile}. The offered GPU services work on a NVidia GeForce GTX 1080 Ti with 11GB memory, CUDA version 10.1, and CUDA driver version 418.67. Pre-processing, launching, and post-processing are done in the JEO-lab layer of the platform in a jupyter notebook docker container, running Tensorflow 1.3.0.

\subsubsection{Equivalent Reference Probability filter}
\label{sec:info_entropy_theory}
Post-processing results from ML/DL exercises is an established practice in practically all such workflows (\citep{gao2022selecting}, \citep{bruha2000postprocessing}, \citep{bansal2020post}). What it usually consists of is the selected removal, based on some criteria, of a substantial enough number of the incorrectly classified examples in order to increase model performance, while simultaneously not falling into the trap of "cherry-picking" one's results. 

In classification problems analysts can employ a filter on probability - keeping only examples for which the network has output a MP of the winning class above a threshold. The analyst then decides where to put the threshold in order to control the rigorousness of the filter - higher for more stringent classification, and lower for a more lenient one. The first problem with this is that it depends heavily on the user's decision and is thus, to a degree, arbitrary. The next problem is that the filter is one dimensional - one can only set a threshold along a single axis. Introducing other, or indeed multiple, dimensions to this process would allow for different spreads of the data in the given space. The intuition is that given the chosen dimensions, the data would neatly split between correct and incorrect classifications and allow for more precise filtering. The desired outcome from such filtering would be to remove the biggest amount of incorrectly classified examples, without removing too many correctly classified ones.

%%original text
% The proposed method works with two fundamental metrics in information theory - information and Shannon entropy in order to produce a scatter-plot of the distribution of values. Information can be defined as the measure of surprise from an event - rare or low probability events are surprising and hence carry more information, and vice versa. It is calculated by taking the negative log of the maximum probability (Equation \ref{eq:info}). Entropy can be defined as calculating the information for a random variable with a given distribution or simply calculating the information for the probability distribution of the events of a given variable (Equation \ref{eq:entropy}). In the present case a low entropy means there is a more pronounced difference between the maximum probability for a given class and the rest of the probabilities for the remaining classes.

%%ERP text
The proposed method works with a metric, based on information theory - the Equivalent Reference Probability (ERP), as described in \citep{bogaert2017information}. In information theory Information is the measure of surprise from an event - rare or low probability events are surprising and hence carry more information, and vice versa (Equation \ref{eq:info}). Entropy is the information for the probability distribution of the events of a given variable (Equation \ref{eq:entropy}). A low entropy means there is a more pronounced difference between the MP for a given class and the rest of the probabilities for the remaining classes. In \citep{bogaert2017information} the authors make use of the difference of information (Equation \ref{eq:edii}) between a reference class (preferably the most probable class) and all the other classes in the probability vector. Because the $E[D(i||i^{*})]$ is unrestricted in terms of potential values, and because it needs an upper and lower band in order to be interpreted, the authors suggest using ERP (Equation \ref{eq:erp}). It is a single metric where $ERP \in \{0,...,1\}$; values approaching 1 mean a very high confidence in the most probable class with the most equal distribution of the remainder to the other $k$ classes.

\begin{equation}
\label{eq:info}
h(x) = -log_2( p(x))
\end{equation}

\begin{equation}
\label{eq:entropy}
H(X) = -\sum_{i=1}^{n} P_i log_2 P_i
\end{equation}

\begin{equation}
\label{eq:edii}
E[D(i||i^{*})] = \log p_i^{*} - \frac{1}{1 - p_i^{*}} \sum_{i \backslash i^{*}} p_i \log p_i
\end{equation}

\begin{equation}
\label{eq:erp}
p^{*} = \frac{exp(E[D(i||i^{*})])}{exp(E[D(i||i^{*})]) + k - 1}
\end{equation}

The appropriate thresholds for ERP and probability are ascertained with a custom function that iteratively moves the threshold down the line. At each step it counts the number of disqualified incorrect images, while trying to keep the number of correctly classified ones below a certain percent. The settable parameter to the function is thus the percentage of correctly classified examples the analyst is willing to discard.
After ascertaining the thresholds the space within the scatter plot is divided into four quadrants. Through the iterative exclusion of one or combinations thereof of the examples in these quadrants, the analyst can perform a more precise filtering on results.  

\section{Results}
\label{sec:results}
Results are divided into five sections. Firstly we present the MMEC dataset, secondly the best performing model is presented, third - the confusion matrix and M-F1 score for best performing model is shown alongside the Producer (PA) and User (UA) Accuracy, fourth the improvement generated from employing an ERP filter, and lastly - we present the performance of the model when faced with images from unfavorable conditions for each class, simulating operational use of the model.

\subsection{Mature Major European Crops}
\label{sec:res_dataset}
\begin{figure*}[!h]
    \centering\includegraphics[width=0.7\linewidth]{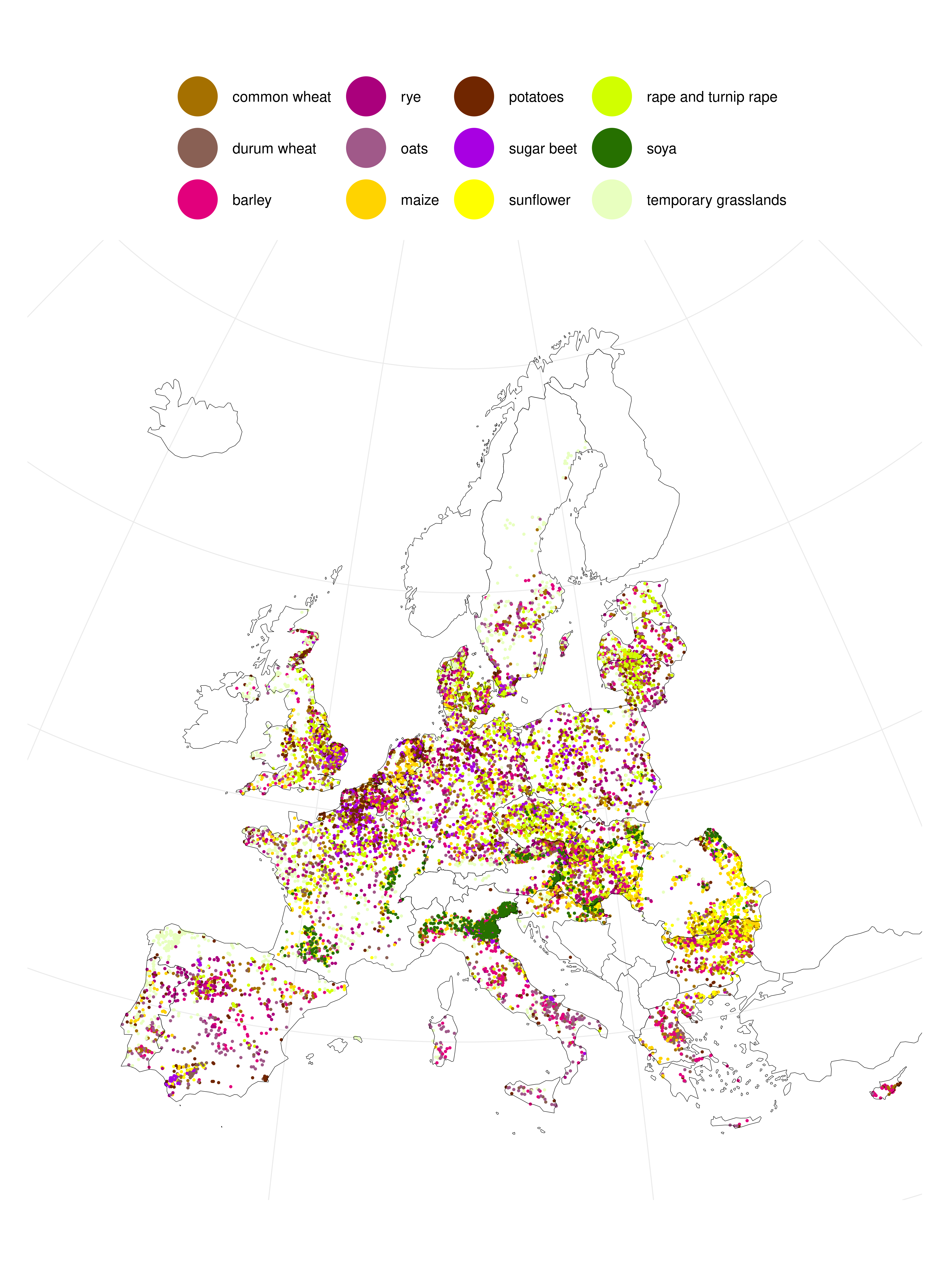}
\caption{Geographical distribution of 15,876 LUCAS Cover photos across the EU, which have been manually screened and validated as ready-to-train-on. Map projection EPSG:3035.}
\label{fig:lucasV_trainok_geog}
\end{figure*}

The processing chain from Sections \ref{sec:ccs} and \ref{sec:experts} produces a dataset of 169,460 LUCAS photos of mature crops across 25 EU Member States. Utilizing the manual labelling as described in Section \ref{sec:manual} the study also publishes 15,876 high quality, ready-to-train-on photos. Each of which has been manually checked and verified to exhibit a clear view to the crop in its mature, pre-harvest stage with no visual obstructions, or foreign objects into the frame. Each class has more than 400 photos, allowing for considerable lee-way in training set selection. A breakdown per country is visible on Table \ref{tab:trainok} and geographical visualization of the same on Figure \ref{fig:lucasV_trainok_geog}.

\begin{table*}[ht]
\footnotesize
\centering
\resizebox{\textwidth}{!}{\begin{tabular}{p{1.15cm}|rrrrrrrrrrrr|r||r}
  \hline
 & \textbf{B11} & \textbf{B12} & \textbf{B13} & \textbf{B14} & \textbf{B15} & \textbf{B16} & \textbf{B21} & \textbf{B22} & \textbf{B31} & \textbf{B32} & \textbf{B33} & \textbf{B55} & \textbf{Total} & \textbf{Total MMEC} \\ 
  \hline
\textbf{AT} & 136 &  32 & 139 & 103 &  29 &  69 &  48 &  67 &  18 &  85 & 122 & 124 & 972 & 3595 \\ 
  \textbf{BE} &  59 &   2 &  71 &   4 &   3 &  39 &  93 &  49 &   0 &  23 &   0 &  62 & 405 & 2127 \\ 
  \textbf{BG} & 179 &   5 & 129 &  19 &  12 & 148 &  11 &   0 & 110 &  90 &   4 &   3 & 710 & 2855 \\ 
  \textbf{CY} &  15 &   1 &  29 &   0 &   2 &   0 &   6 &   0 &   0 &   0 &   0 &   0 &  53 & 207 \\ 
  \textbf{CZ} &  72 &   4 &  28 &  32 &  20 &  47 &  30 &  45 &   4 & 194 &  12 &  55 & 543 & 6691 \\ 
  \textbf{DE} & 156 &  40 & 157 & 176 & 133 & 114 & 177 & 152 &  17 & 220 &   6 & 212 & 1560 & 24055 \\ 
  \textbf{DK} & 132 &   2 &  93 & 105 &  27 &  55 &  21 &  24 &   0 & 184 &   0 & 175 & 818 & 3226 \\ 
  \textbf{EE} &  20 &   0 &  16 &   7 &   8 &   1 &   5 &   0 &   0 &  26 &   0 &  14 &  97 & 612 \\ 
  \textbf{EL} &  62 &  81 &  88 &   0 &  25 &  62 &   6 &   7 &  22 &   9 &   0 &   4 & 366 & 1386 \\ 
  \textbf{ES} &  68 &  58 &  85 & 105 &  80 &  22 &  59 &  58 &  34 &  50 &   0 & 178 & 797 & 19582 \\ 
  \textbf{FR} & 121 &  97 & 116 & 130 & 118 &  82 & 139 & 143 &  75 & 186 & 142 & 193 & 1542 & 40989 \\ 
  \textbf{HR} &  38 &   0 &  22 &   4 &   7 &  53 &   6 &   2 &  16 &   9 &  34 &  16 & 207 & 434 \\ 
  \textbf{HU} &  74 &  34 & 103 &  66 &  30 &  53 &  17 &   8 &  49 & 135 &  49 &   8 & 626 & 7354 \\ 
  \textbf{IT} & 136 &  54 & 175 &  17 & 130 &  54 &  50 &  98 &  17 &  21 & 378 & 166 & 1296 & 13387 \\ 
  \textbf{LT} &  82 &   6 &  65 &  72 &  50 &   1 &  28 &   4 &   0 & 147 &   0 &  31 & 486 & 2313 \\ 
  \textbf{LU} &   3 &   0 &   7 &   2 &   0 &   1 &   0 &   0 &   0 &   1 &   0 &   6 &  20 & 149 \\ 
  \textbf{LV} &  66 &   3 &  72 &  44 &  33 &   9 &  12 &   1 &   0 & 110 &   0 &  42 & 392 & 1763 \\ 
  \textbf{NL} & 150 &   0 &  53 &  23 &   2 & 134 & 200 & 115 &   0 &   3 &   0 &  54 & 734 & 1805 \\ 
  \textbf{PL} &  66 &  12 &  40 & 105 &  61 &  98 &  89 &  93 &   1 & 173 &   7 &  67 & 812 & 20542 \\ 
  \textbf{PT} &  28 &   7 &  24 &  40 &  50 &  22 &  19 &   0 &   1 &   0 &   0 &  68 & 259 & 877 \\ 
  \textbf{RO} &  71 &  28 &  39 &  13 &  25 & 159 &  22 &  16 & 189 &  51 &  88 &  32 & 733 & 3649 \\ 
  \textbf{SE} &  89 &   0 &  67 &  47 &  85 &  11 &  34 &  55 &   0 &  95 &   0 & 167 & 650 & 2742 \\ 
  \textbf{SI} &  29 &   6 &  26 &   2 &   2 &  30 &   6 &   0 &   1 &   5 &   0 &  27 & 134 & 364 \\ 
  \textbf{SK} &  89 &  13 &  95 &  27 &  16 &  51 &  18 &  25 &  42 & 164 &  55 &  18 & 613 & 3091 \\ 
  \textbf{UK} & 155 &   0 & 134 &   8 &  84 &  78 & 126 & 105 &   0 & 182 &   0 & 179 & 1051 & 5665 \\ 
    \hline
  \textbf{Total} & 2096 & 485 & 1873 & 1151 & 1032 & 1393 & 1222 & 1067 & 596 & 2163 & 897 & 1901 & 15876 &   - \\ 
    \hline
      \hline
  \textbf{Total MMEC} & 47143 & 8062 & 31500 & 7296 & 6582 & 32175 & 4113 & 4414 & 6830 & 13958 & 1603 & 5784 &   - & 169460 \\ 
\end{tabular}}
\caption{All visually inspected MMEC photos labelled as good to train on from the manual annotation using PyGeon. The marginal rows labelled \textit{MMEC} show the total number of photos that show mature crops, which have not been visually inspected.}
\label{tab:trainok}
\end{table*}

\subsection{Best performing model}
The models were ranked using OA on an independent inference set of 85 images per class (Table \ref{tab:top3augmentedPlusBest}). The best model was identified as number 78, achieving an OA of 79.4\%. The relevant parameter settings are - Learning Rate of 0.0035148759, Batch size of 1024, Momentum of zero, as the Optimizer used is Gradient Descent. The M-F1 and OA on the test set are identical, as we are dealing with a balanced inference set. The last column shows the best model (78), applied over the imbalanced inference set (see Section \ref{sec:sets}). The model is exposed to 7722 more examples, and the drop in M-F1 is 0.0369, meaning the model is trained and generalizes very well over larger datasets.

\begin{table}[ht]
\centering
\footnotesize

\caption{Output for the top three performing models with augmentations plus the output from the best model ran on the imbalanced set. The applied augmentation were left-right flip and random brightness. The table shows (in order) the model number, along with the relevant configuration (Learning rate, Batch size, Momentum, Optimizer), the number of labelled images, the training and validation accuracy, and the M-F1.}
\label{tab:top3augmentedPlusBest}
\begin{tabular}{r|llll}
  \hline
\textbf{Ranking} & \textbf{1} & \textbf{2} & \textbf{3} & \textbf{Best} \\ 
  \hline
Model & 78 & 88 & 4 & 78 \\ 
  Level & Augm & Augm & Augm & Best Model \\ 
  LR & 0.0035 & 0.0073 & 0.0096 & 0.0035 \\ 
  BS & 1024 & 512 & 512 & 1024 \\ 
  Momentum & 0 & 0 & 0 & 0 \\ 
  Optimizer & GD & GD & GD & GD \\ 
  \# of Images & 1020 & 1020 & 1020 & 8642 \\
  \hline
  Validation Accuracy & 0.7768 & 0.7789 & 0.7747 & 0.7768 \\ 
  Training Accuracy & 0.8945 & 0.8965 & 0.9238 & 0.8945 \\ 
  Test Accuracy & 0.7941 & 0.7775 & 0.7755 & 0.7854 \\ 
  M-F1 & 0.7941 & 0.7775 & 0.7755 & 0.7572 \\ 
   \hline
\end{tabular}
\end{table}

\subsection{Confusion Matrix}
\begin{figure*}[!h]
    \centering\includegraphics[width=0.8\linewidth]{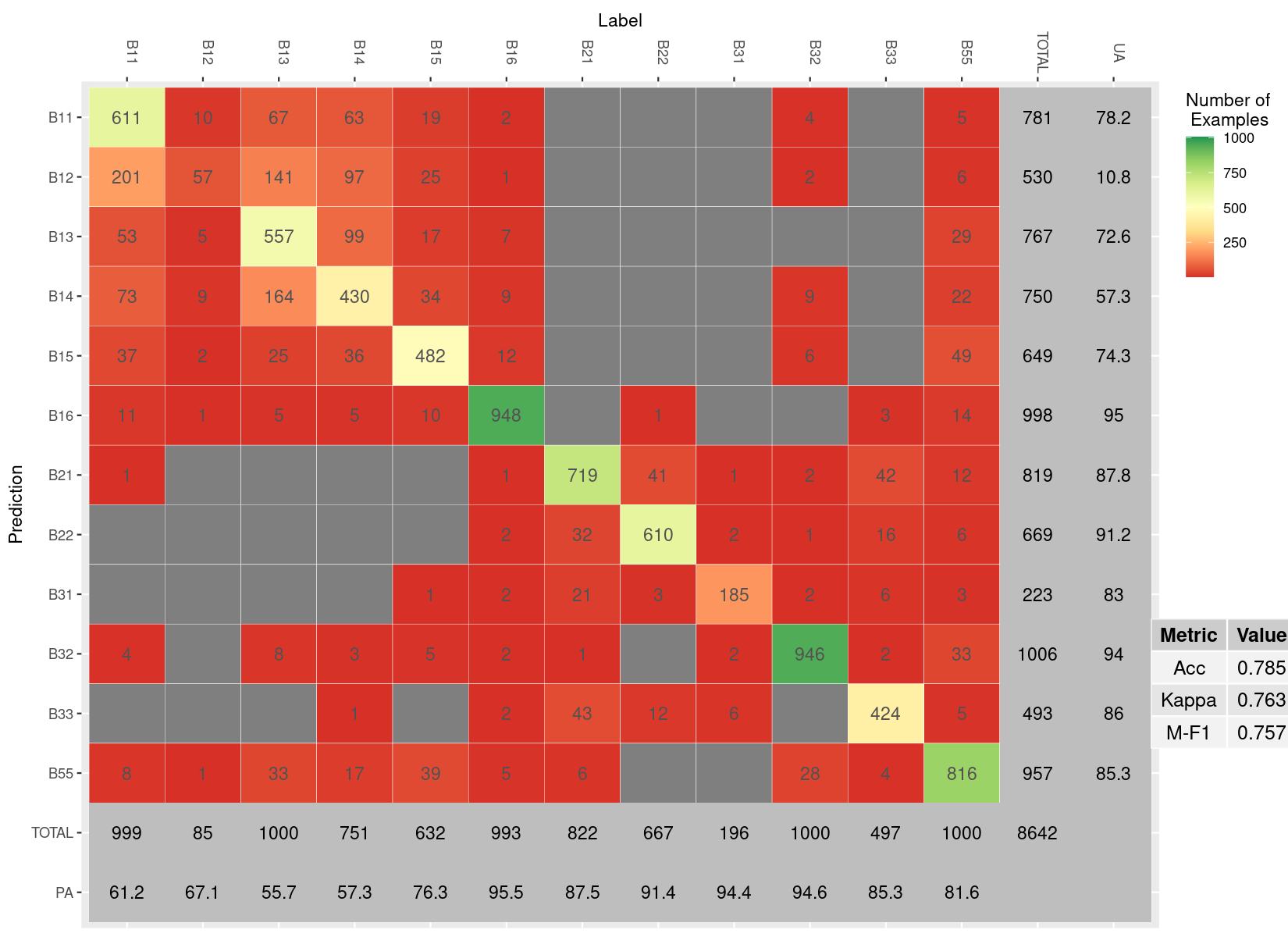}
\caption{Confusion matrix for best model (78) over imbalanced inference set.}
\label{fig:confmatr}
\end{figure*}

The confusion matrix for the best model run (78) over the imbalanced operational inference set is presented in Figure \ref{fig:confmatr}. It is clear that the majority of confusion happens between the cereal classes (B11-B15) and with Grassland (B55). In fact, the difference between the average PA of all crops, excluding Grassland, and the average PA of the cereal classes is 27.9, and for UA the difference is 30.9. The class which gets most commonly miss-classified as a false positive is Durum wheat (B12) with a UA of 10.8; the low score arguably has much to do with the unequal representation of the class. The best performing class is Maize (B16), with a PA of 95.5 and UA of 95, followed closely by Rape and turnip rape (B32), showing the clear separation of both from the other classes.

\subsection{Equivalent reference probability filter}
\begin{figure*}[!h]
    \centering\includegraphics[width=\linewidth]{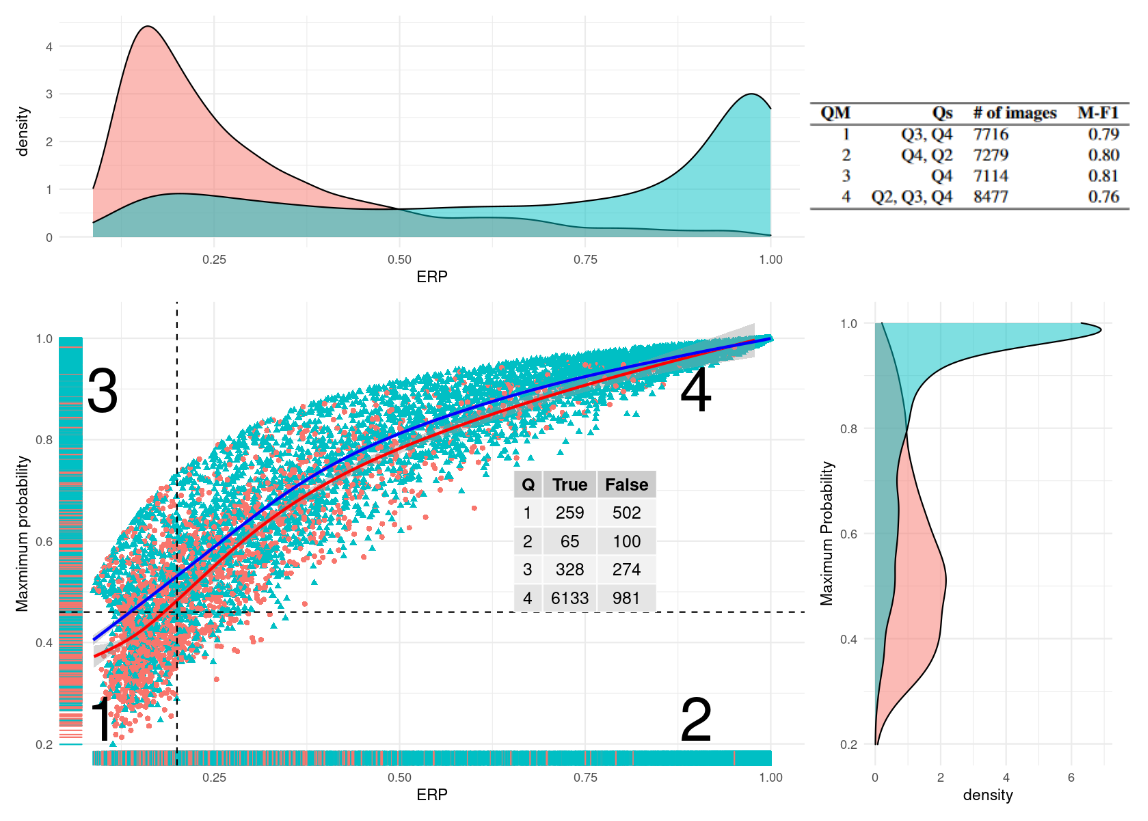}
\caption{Scatter plot of ERP and probability quadrant filtering with marginal density plots for each variable. In red are the incorrect and in blue - the correct classifications. Numbers within quadrants indicate the quadrant ID. The data is fitted with smooth lines for correct and incorrect classifications. The table in the uppermost corner shows the results from the quadrant filtering. In order the columns represent - the quadrant method ID, the quadrants included in the method, the number of images, and the M-F1 achieved through the inclusion of the respective Qs. In order the QMs represent -  1. MP only, 2. ERP only, 3. both above their resp. thresholds, 4. at least one above its threshold.}
\label{fig:entropyinfo}
\end{figure*}

The application of the quadrant filtering method using ERP and MP is shown in Figure \ref{fig:entropyinfo}. The dotted lines represent the thresholds identified by the functions described in Section \ref{sec:info_entropy_theory}. The settable parameter is fixed at loosing no more than one percent of the correctly classified images, meaning the identified thresholds are the most conservative ones. They are 0.46 for MP and 0.2 for ERP. The inscribed table shows the number of true and false classifications in each quadrant as labelled by their respective quadrant ID. Although similar, their is a notable difference in the distribution of the true and false classifications, visible in the smooth fitted lines for each group.

%original with entropy
% \begin{figure*}[!h]
%     \centering\includegraphics[width=\linewidth]{figures/entropyAndProbaWMarginals_78.png}
% \caption{Scatter plot of Information and Entropy quadrant filtering with marginal density plots for each variable. Numbers within quadrants indicate the quadrant ID. The data is fitted with smooth lines for each group.}
% \label{fig:entropyinfo}
% \end{figure*}

The results achieved from employing such filtering are presented in Figure \ref{fig:entropyinfo} in the table in the uppermost right corner. There is an M-F1 increase of 0.6 from not using any filter and of 0.2 from using only the MP filter.

%%% original with entropy
% \begin{table}[ht]
% \centering
% \begin{tabular}{rrlrr}
%   \hline
% \textbf{QM} & \textbf{Qs} & \textbf{\# of images} & \textbf{M-F1} \\ 
%   \hline
% 1 & Q1, Q2 & 7415 & 0.803 \\ 
% 2 & Q1, Q3 & 7211 & 0.809 \\ 
% 3 & Q1 & 7032 & 0.817 \\ 
% 4 & Q1, Q2, Q3 & 7594 & 0.796 \\ 
%   \hline
% \end{tabular}
% \caption{Results from quadrant filtering. In order the columns represent - the quadrant method ID, the quadrants included in the method, the number of images, and the M-F1 achieved through the inclusion respective QM.}
% \label{tab:QMs

%with ERP
% \begin{table}[ht]
% \centering
% \begin{tabular}{rrlrr}
%   \hline
% \textbf{QM} & \textbf{Qs} & \textbf{\# of images} & \textbf{M-F1} \\ 
%   \hline
% 1 & Q3, Q4 & 7716 & 0.79 \\ 
% 2 & Q4, Q2 & 7279 & 0.80 \\ 
% 3 & Q4 & 7114 & 0.81 \\ 
% 4 & Q2, Q3, Q4 & 8477 & 0.76 \\ 
%   \hline
% \end{tabular}
% \caption{Results from quadrant filtering. In order the columns represent - the quadrant method ID, the quadrants included in the method, the number of images, and the M-F1 achieved through the inclusion respective QM.}
% \label{tab:QMs}
% \end{table}

\subsection{Unfavorable conditions}
\begin{figure*}[!h]
    \centering\includegraphics[width=0.7\linewidth]{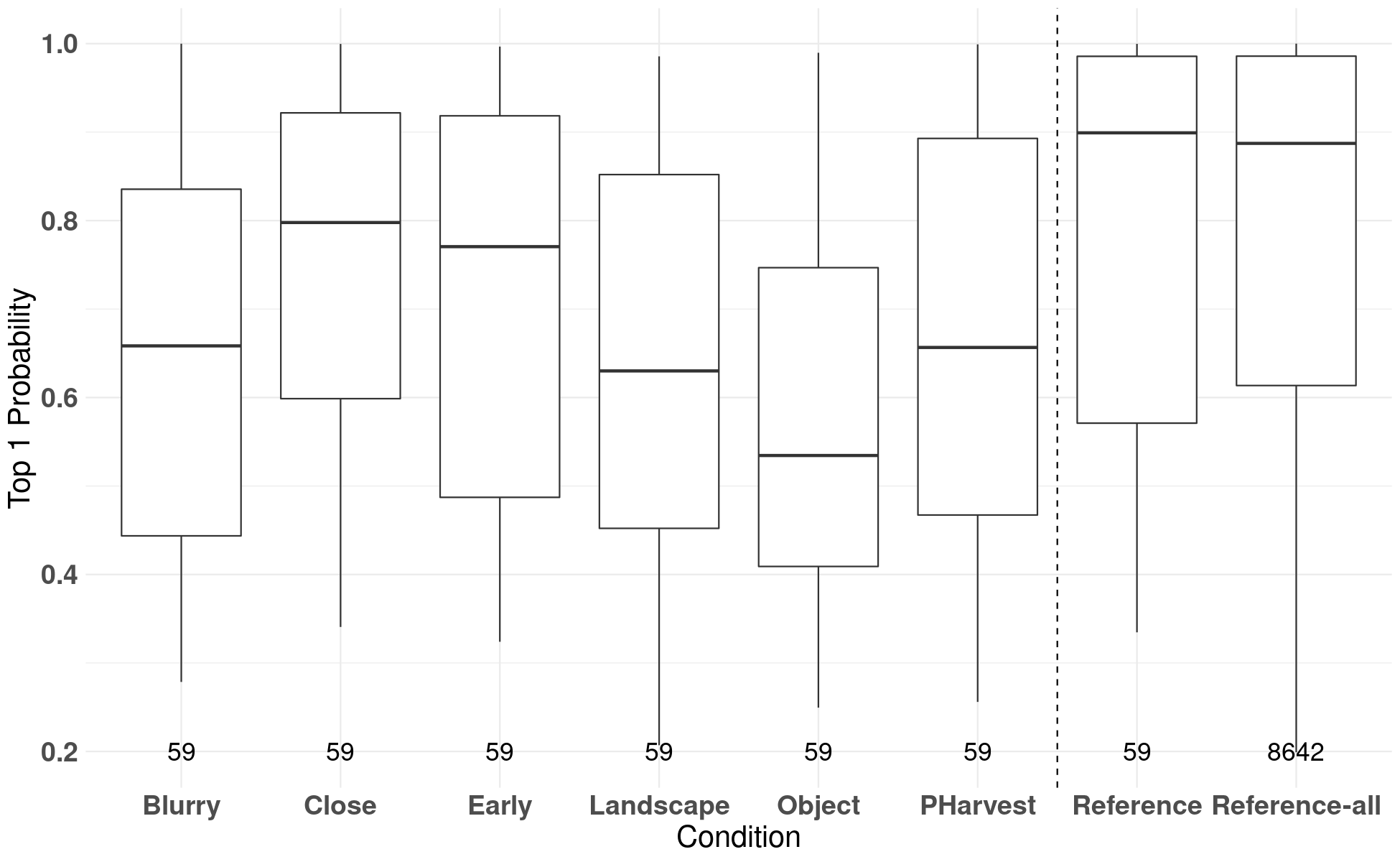}
\caption{Top1 probability for all examples for a given unfavorable condition with reference to random sample of the same size of the balanced inference set and to the entire imbalanced inference set. Number of examples in each box is given above the condition label.}
\label{fig:boxplotbadimgs}
\end{figure*}

Best model 78 was applied over a stratified sample of 1 photo per year, per LUCAS LC1 class, and per unfavorable condition, totalling at an inference set of 354, meaning 59 photos per unfavorable condition (see the examples in Figure \ref{fig:badimgs}). A boxplot of the Top1 probability for each unfavorable condition is presented on Figure \ref{fig:boxplotbadimgs}. The conditions are compared firstly to a reference set of quality images that are randomly sampled to have the same distribution as the sets of the conditions, and secondly to the entire imbalanced inference set. Model 78 is most confused about photos with foreign objects, landscape photos, and photos showing the crop post its harvest period, with blurry, early and especially close-up photos performing significantly closer to the reference in terms of Top1 probability. 

The actual classification results are presented on Table \ref{tab:unf_con_tab}. The worst results are achieved with photos exhibiting post-harvest conditions with an OA of 20\%, and early and examples with a foreign object in the frame following with 31\% and 37\% respectively. The unfavorable conditions that impact the performance the least are blurry and overly close-up photos (54\%). This illustrates that a clear protocol is needed when such automated procedures are used within operational workflows, such as for the CAP \cite{sima2020use}. In addition, models can progressively be trained with a set of photos covering a wider range of conditions to improve their generalization capacity.

\begin{table}[ht]
\centering
\begin{tabular}{rrrr}
  \hline
 & False & True & OA \\ 
  \hline
Blurry &  27 &  32 & 0.54 \\ 
  Close &  27 &  32 & 0.54 \\ 
  Early &  41 &  18 & 0.31 \\ 
  Landscape &  35 &  24 & 0.41 \\ 
  Object &  37 &  22 & 0.37 \\ 
  Post-harvest &  47 &  12 & 0.20 \\ 
   \hline
\end{tabular}
\caption{Number of true and false classifications and overall accuracy for each unfavorable condition.}
\label{tab:unf_con_tab}
\end{table}

%Figure 1. Example pictures of selected LUCAS cover pictures of mature crops

%Figure 2. Map distribution of LUCAS points with cover pictures of mature crops in the full dataset. 

%Figure 3. 

%Confusion matrices
%Spatial patterns

\section{Discussion}
\subsection{Context}

%%literature comparison and place
Recently, several relevant studies were published. \citet{zheng2019cropdeep} present the CropDeep dataset, over which they test state of the art classification and detection DL algorithms. They achieve an averaged accuracy of 99.81\% over the CropDeep datasets. These results are impressive, although not directly comparable, as the images were collected from robots in a sterile greenhouse environment, allowing for image conditions to be identical between acquisitions. They furthermore used average accuracy as a metric over an imbalanced inference set, which is not in accordance with the literature \citep{ma2013imbalanced}. \citet{gao2021identification} achieved an accuracy of 99.51\% in differentiating 30 wheat cultivars at the flowering (most mature) stage. This is very impressive, considering the present study suffered the most error when trying to discriminate between the various cereal classes. The difference is again in the lab quality of the images taken, whereby each image exhibits a single plant on a white background. \citet{d2022monitoring} achieved a M-F1 score of 62.3\% for 10 classes using street level images. The current study outperformed the cited work by 13.4\%, though this can be attributed to the lower presence of noise on the images fed to the model. 

%% compare with pl@ntNet results here

%% novelties
This study presents the first use of the LUCAS cover dataset for automatic crop identification.  Indeed, it is the first study to apply DL for crop identification on still images that are not taken in a controlled environment and coming from a wide variety of sensors, which truly mimics an operational scenario. Secondly, the study produces an automated way to attach crop life-cycle stage information to a database of photos. Third, the introduction of quadrant filtering is a step towards a new State of the Art for more precise post-processing filtering. Whether using crop calendars to extract photos for specific crop life-cycle stages, or using the dataset as a whole, the authors belief that various lines of research may be developed using the LUCAS cover photos. 

\subsection{ERP filtering}
%some context
A main achievement of the study is the exploration of methods for filtering classification results to achieve better performance and to quantify uncertainty. The study made use of ERP as a metric for assessing this uncertainty. According to the literature, ERP has been shown to be more robust than MP in classifying pixel-level thematic uncertainty \citep{bogaert2017information}; more precise than majority voting in post-processing speckle removal of classified maps \citep{waldner2017national}; and more flexible than OA in terms of independence of the distribution of the validation data \citep{owusu2021towards}. 

\begin{figure*}[!h]
    \centering\includegraphics[width=0.8\linewidth]{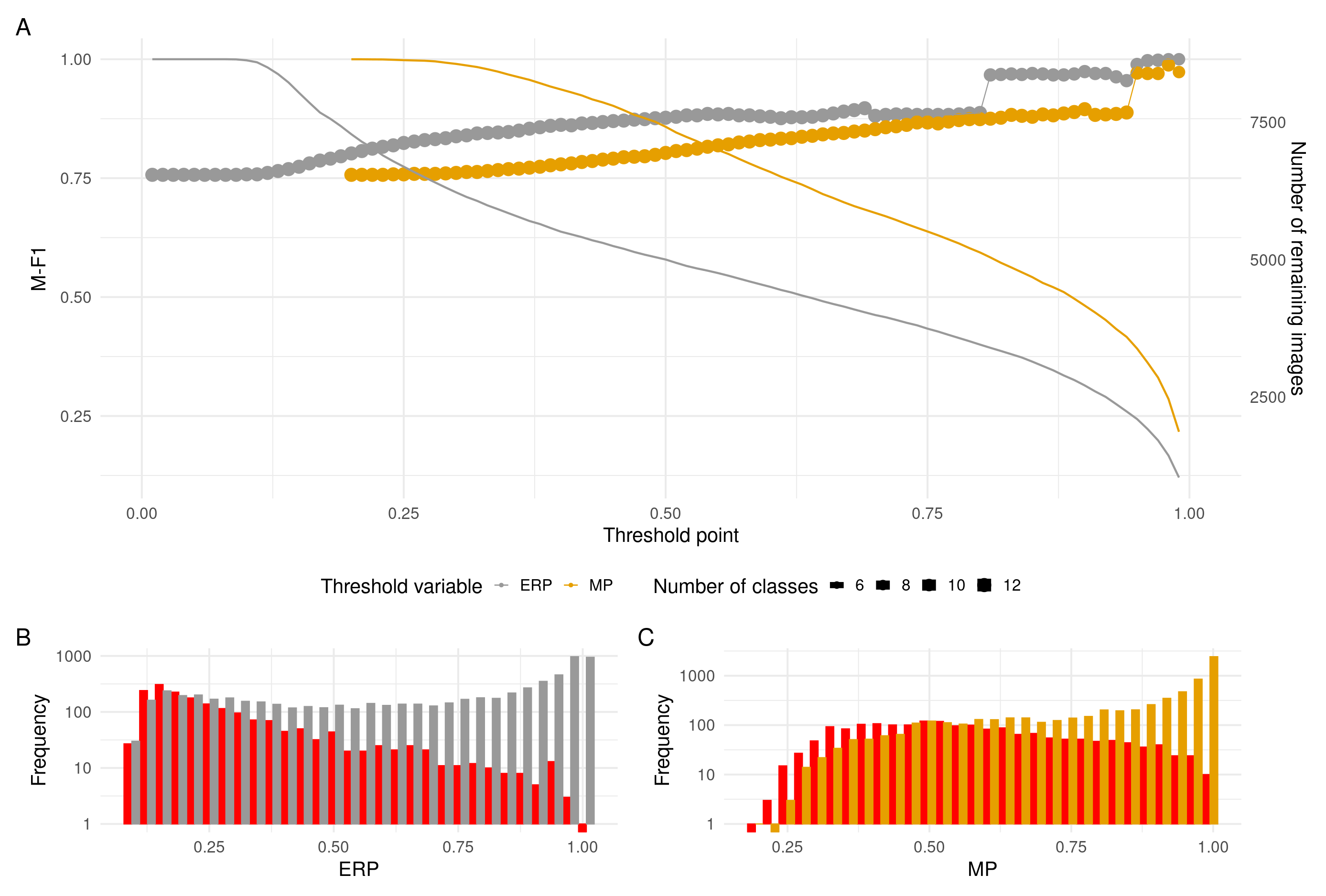}
\caption{Comparison between using MP and ERP for applying filtering on results. Plot A's first Y axis shows the evolution in M-F1 in connected points when applying a threshold on the inference set; and it's secondary Y axis - the diminishing number of photos in continuous planes when applying the same thresholds. Plots B and C represent the distributions in logarithmic scale (base 10) of values present in the inference set in terms of ERP and MP respectively. The red bars in both histograms are the incorrect classifications.}
\label{fig:proba_erp_evol}
\end{figure*}

%fig 9, subplot A
In practice, MP and ERP are connected, which is clearly visible in the distribution of both groups (correct and incorrect) in the space where the joint probability reference distribution is not null on Figure \ref{fig:entropyinfo}. From the marginal distribution plots we can see that this connection is inverted - there is a high peak in the low values of MP for the incorrectly classified points and a high peak in the high values of ERP for the correctly classified ones.
Furthermore, as shown in Figure \ref{fig:proba_erp_evol}, ERP performs significantly better than MP in post-processing filtering. Because ERP and MP are both probabilities that are in the range between zero and one, their direct comparison in this regard is straightforward. Firstly in subplot A, where for an equal threshold value, the M-F1 value is always higher when utilising ERP over MP. This means that ERP is a much better estimator of uncertainty and manages to capture to a finer degree the nuances that distinguish an incorrect from a correct classification. It needs to be mentioned that this is partly due to the fact that, while for MP the smallest possible threshold value is relatively high (0.20), with ERP it is found at the first stage of filtering (0.01). As seen on the secondary Y axis, which shows the number of images left in the set after performing the filter, this process is not without cost - the number is, for every threshold value, less for ERP than for MP. Nevertheless, it is always preferable to have a bigger spread of the data, over which to set thresholds, especially so when the analysis needs to be conservative regarding the number of correct classification it is willing to lose.

%fig 9 histograms
Furthermore, the histograms in subplots B and C show the point at which the proportion of correct and incorrect classifications for each threshold value, represented by the height of the gray and yellow bars, relative to the red bar, changes in favour of the correct ones. While with MP this point arrives at 0.54, for ERP the change is present already at 0.24. Hence the relative cost in terms of number of examples disqualified due to the threshold setting is proportionately lower with ERP in order to achieve the same increase of M-F1.

\subsection{Limitations}

%% crop calendar information collected from different sources + expert gap filling

Although several novel aspects have been highlighted, some limitations are present in our study. Firstly, there are issues with the pre-processing of the data - in particular, the fact that CC information comes from a variety of sources. Albeit official CCs, which have been harmonized, the fact that the study is treating them as a-priori semantically harmonized could be problematic. Because organizations, based on their goal, have different data collection, processing, and publishing protocols, it is conceivable that the data was intended for a different use. The issue becomes even more apparent when considering the expert knowledge and model output gap filling. Indeed, the concern for the latter introducing error into the results was such, that the study went ahead and calculated the M-F1 for each country (NUTS0 region), for which the crop calendar information was derived from expert knowledge or model output, and then compared to the reference M-F1. No clear drop in M-F1 based on the origin of the mature crop information was registered from this analysis.

%% introducing bias through manual photo selection with PYgeon + for images with unfavorable conditions

Another data issue is that bias can be introduced during manual selection by visual assessment. Other than errors due to distraction during annotation, the annotator undoubtedly bases their decision on which images to keep and discard based on their own discretion. For example, the annotator had to consider questions like - should there be any sky or abundance of soil visible on the image; is the crop on this image to be considered mature enough; and, especially so for the cereal classes - is this the correct label. The matter is even more pronounced when selecting examples for unfavorable conditions. During which, for example, the distinction between "Blurry" and "Close" was sometimes hard to make, the object in the "Object" class and the visual appearance of the landscape in the "Landscape" one were very varied, and that sometimes the image showed more than a single unfavorable condition - the crop can be both early in the season and blurred out, in which case one could have used multi-tags. Such issues were considered prior to undertaking each task, yet the possibility of bias has to be mentioned.

%% information/entropy cutoff point function - already looking into correct-incorrect classifications to arrive at the cutoff - cart before the horse. ideally one would let the values of the variables that arise naturally from the data to guide the decision-making, rather than directing the process through "peeking" in the result of the classification in order to make a decision about disqualifying examples.

Secondly, there are issues related to the processing logic of certain steps. One such is the identification of threshold points for MP and ERP to generate quadrants. The way the custom function works is by peeking into the correct-incorrect classification results in order to iteratively arrive at the threshold with the main consideration being keeping the number of disqualified correct classifications below a certain percentage. In a sense this means putting the proverbial data cart before the horse, as instead of using simply the values of whichever chosen metrics, the function also considers the result of the classification.

\subsection{Recommendations}
\label{sec:recommendations}

%% consider grouping cereals into a common class and if one wants to distinguish betwee the type of cereal - have a separate model for that, as they are too simialr - show the diff in performance if cereals were a common class
There are several recommendations that would be a logical continuation of the work. In terms of class selection, the major part of the confusion stems from the cereal classes (Section \ref{sec:results}). This makes sense, as to distinguish between them can sometimes be troublesome even for a skilled professional. As a grouped cereal class they are easily set apart from the rest of the crops, but between them, the structure of the fruit, stem, and leaf organs can look too similar. Indeed, the approach in \citet{gao2021identification} yields such good results exactly because the model is designed to pick up on the subtle differences between the varieties. In the present case grouping the cereals together would produce a M-F1 of 88.2 without and 90.4 with quadrant filtering, which is 12.5 and 14.7 points higher than the achieved result. Ideally one would capture the cereal class first at these higher ranges of M-F1 and then have a separate model that deals solely with classifying the type of cereal, variety, or cultivar.

%% use different metrics from information theory to form other orthogonal spaces or indeed spaces of higher dimensionality
Concerning the point of being more robust in identifying thresholds, or more generally on the topic of splitting the space in Figure \ref{fig:entropyinfo}, one could build a kind of Bayesian Discriminant Rule in order to generalize the combination of the two 1-D thresholds to a 2-D threshold. This can be done by taking into consideration the joint distributions and would yield a single curve that separates correctly and incorrectly classified examples.
%The introduction of another dimension to form orthogonal spaces when performing post-processing filtering is undoubtedly worthwhile. The metrics used in this study are the basic notions from information theory of Shannon information and entropy. In reality, there are many ways to examine the behavior of large sets of random variables and their probability distributions, such as the output of DL exercises, using concepts from branches of statistics such as probability and information theory. For example, one might do a pair-wise comparison of the separate classes using mutual information and joint entropy in order to try and estimate the information and entropy relationships between the various classes. 

%% effect of imamgeSize on top1 prob and cnn_result
%% use different nets of different inputSize
An always current topic in DL for CV is the effect of resolution on results. In this case, one can discuss both the input resolution of the source images, and the input resolution of the net in use. Firstly, the range of values of images' resolutions in the inference set vary between 480-3504 in height and 640-4672 in width - a 7.3 times difference in each dimension. Almost 65\% of the images are of resolution 1600x1200, with another 22\% being 2048x1563 (for full breakdown of available image resolutions check Supplementary Table \ref{tab:resolultions}). With such a spread one can imagine that the level of detail visible on images from either end of the range is quite different. When measuring the correlation between image resolution and the proportion of correctly classified examples for each resolution bin (Figure \ref{fig:lucasv_wxh_propr_point}), the study found an R-squared value of 0.009, meaning the correlation for this set of LUCAS photos is almost none. Secondly, the net input size is 224x224, meaning each parallelogram image of the training and inference set gets re-scaled to this square size. Intuitively, one can say that larger images would lose more information during re-scaling than smaller ones. In reality, the re-scaling turns the problem into a detection of the major structural features of the crops (e.g. broad leaf vs cereals, colouring, having recognisable flowers or not), where resolution does not matter as much. This would also shed light as to the reason why the network has trouble distinguishing between cereal classes. The analysis still serves to illustrate that the method is developed to handle equally well images from different resolutions. This further showcases the policy relevance of the work, as in an operational context, a regulating body is expected to receive evidence-images in a variety of image resolutions.

\section{Conclusion}
This study provides a subset of LUCAS Cover photos for 12 major crops across the EU, to deploy, benchmark, and identify the best configuration of Mobile-net for the classification task, to showcase the possibility of using entropy-based metrics for post-processing of results, and finally to show the applications and limitations of the model in a practical and policy relevant context.  The work has produced a dataset of 169,460 images of mature crops for the 12 classes, out of which 15,876 were manually selected as representing a clean sample without any foreign objects or unfavorable conditions. The best performing model to identify crop achieved a Macro F1 (M-F1) of 0.75 on an imbalanced test dataset of 8,642 photos. Using metrics from information theory resulted in achieving an increase of 6\%. The most unfavorable conditions for taking such images, across all crop classes, were found to be to early or late in the season. The proposed methodology shows the possibility for using minimal auxiliary data, outside the images themselves, in order to achieve a M-F1 of 0.817 for labelling between 12 major European crops. 

\bibliographystyle{plainnat}
\bibliography{sample.bib}

\begin{thebibliography}{44}
\providecommand{\natexlab}[1]{#1}
\providecommand{\url}[1]{\texttt{#1}}
\expandafter\ifx\csname urlstyle\endcsname\relax
  \providecommand{\doi}[1]{doi: #1}\else
  \providecommand{\doi}{doi: \begingroup \urlstyle{rm}\Url}\fi

\bibitem[agr()]{agri4castBullArchive}
{AGRI4CAST Bulletin Archive 2018}.
\newblock \url{https://agri4cast.jrc.ec.europa.eu/BulletinsArchive#2018}.
\newblock Accessed: 2021-06-30.

\bibitem[Affouard et~al.(2017)Affouard, Go{\"e}au, Bonnet, Lombardo, and
  Joly]{affouard2017pl}
Antoine Affouard, Herv{\'e} Go{\"e}au, Pierre Bonnet, Jean-Christophe Lombardo,
  and Alexis Joly.
\newblock Pl@ ntnet app in the era of deep learning.
\newblock In \emph{ICLR: International Conference on Learning Representations},
  2017.

\bibitem[AGRI4CAST()]{agri4cast}
AGRI4CAST.
\newblock Joint research center agri4cast resource portal.
\newblock URL
  \url{https://agri4cast.jrc.ec.europa.eu/DataPortal/Index.aspx?o=}.

\bibitem[Alzubaidi et~al.(2021)Alzubaidi, Zhang, Humaidi, Al-Dujaili, Duan,
  Al-Shamma, Santamar{\'\i}a, Fadhel, Al-Amidie, and
  Farhan]{alzubaidi2021review}
Laith Alzubaidi, Jinglan Zhang, Amjad~J Humaidi, Ayad Al-Dujaili, Ye~Duan,
  Omran Al-Shamma, Jos{\'e} Santamar{\'\i}a, Mohammed~A Fadhel, Muthana
  Al-Amidie, and Laith Farhan.
\newblock Review of deep learning: Concepts, cnn architectures, challenges,
  applications, future directions.
\newblock \emph{Journal of big Data}, 8\penalty0 (1):\penalty0 1--74, 2021.

\bibitem[Bansal and Kumar(2020)]{bansal2020post}
Subodh Bansal and Anuj Kumar.
\newblock A post-processing fusion framework for deep learning models for crop
  disease detection.
\newblock In \emph{IOP Conference Series: Materials Science and Engineering},
  volume 998, page 012065. IOP Publishing, 2020.

\bibitem[Bergstra and Bengio(2012)]{bergstra2012random}
James Bergstra and Yoshua Bengio.
\newblock Random search for hyper-parameter optimization.
\newblock \emph{Journal of machine learning research}, 13\penalty0 (2), 2012.

\bibitem[Bogaert et~al.(2017)Bogaert, Waldner, and
  Defourny]{bogaert2017information}
Patrick Bogaert, Fran{\c{c}}ois Waldner, and Pierre Defourny.
\newblock An information-based criterion to measure pixel-level thematic
  uncertainty in land cover classifications.
\newblock \emph{Stochastic Environmental Research and Risk Assessment},
  31\penalty0 (9):\penalty0 2297--2312, 2017.

\bibitem[Bruha and Famili(2000)]{bruha2000postprocessing}
Ivan Bruha and A~Famili.
\newblock Postprocessing in machine learning and data mining.
\newblock \emph{ACM SIGKDD Explorations Newsletter}, 2\penalty0 (2):\penalty0
  110--114, 2000.

\bibitem[Columbus()]{louiscolumbus}
Louis Columbus.
\newblock 10 ways ai has the potential to improve agriculture in 2021.
\newblock URL
  \url{https://www.forbes.com/sites/louiscolumbus/2021/02/17/10-ways-ai-has-the-potential-to-improve-agriculture-in-2021/?sh=316edd747f3b}.

\bibitem[Commission()]{caphomepage}
European Commission.
\newblock The common agricultural policy at a glance.
\newblock URL
  \url{https://ec.europa.eu/info/food-farming-fisheries/key-policies/common-agricultural-policy/cap-glance_en}.

\bibitem[d'Andrimont et~al.(2021)d'Andrimont, Verhegghen, Meroni, Lemoine,
  Strobl, Eiselt, Yordanov, Martinez-Sanchez, and van~der Velde]{d2021lucas}
Rapha{\"e}l d'Andrimont, Astrid Verhegghen, Michele Meroni, Guido Lemoine,
  Peter Strobl, Beatrice Eiselt, Momchil Yordanov, Laura Martinez-Sanchez, and
  Marijn van~der Velde.
\newblock Lucas copernicus 2018: Earth-observation-relevant in situ data on
  land cover and use throughout the european union.
\newblock \emph{Earth System Science Data}, 13\penalty0 (3):\penalty0
  1119--1133, 2021.

\bibitem[d'Andrimont et~al.(2022)d'Andrimont, Yordanov, Martinez-Sanchez, Haub,
  Buck, Haub, Eiselt, and van~der Velde]{d2022lucas}
Rapha{\"e}l d'Andrimont, Momchil Yordanov, Laura Martinez-Sanchez, Peter Haub,
  Oliver Buck, Carsten Haub, Beatrice Eiselt, and Marijn van~der Velde.
\newblock Lucas cover photos 2006--2018 over the eu: 874,646 spatially
  distributed geo-tagged close-up photos with land cover and plant species
  label.
\newblock \emph{Earth System Science Data Discussions}, pages 1--14, 2022.

\bibitem[Devos et~al.()Devos, Sima, and Milenov]{devosconceptual}
Wim Devos, Aleksandra Sima, and Pavel Milenov.
\newblock Conceptual basis of checks by monitoring.

\bibitem[d’Andrimont et~al.(2022)d’Andrimont, Yordanov, Martinez-Sanchez,
  and van~der Velde]{d2022monitoring}
Rapha{\"e}l d’Andrimont, Momchil Yordanov, Laura Martinez-Sanchez, and Marijn
  van~der Velde.
\newblock Monitoring crop phenology with street-level imagery using computer
  vision.
\newblock \emph{Computers and Electronics in Agriculture}, 196:\penalty0
  106866, 2022.

\bibitem[EUROPABIO()]{europabio}
EUROPABIO.
\newblock Europabio official website.
\newblock URL \url{https://www.europabio.org/}.

\bibitem[Gao et~al.(2021)Gao, Liu, Han, Lu, Wang, Zhang, Bai, and
  Luo]{gao2021identification}
Jiameng Gao, Chengzhong Liu, Junying Han, Qinglin Lu, Hengxing Wang, Jianhua
  Zhang, Xuguang Bai, and Jiake Luo.
\newblock Identification method of wheat cultivars by using a convolutional
  neural network combined with images of multiple growth periods of wheat.
\newblock \emph{Symmetry}, 13\penalty0 (11):\penalty0 2012, 2021.

\bibitem[Gao et~al.(2022)Gao, Ram, Philip, Rodr{\'\i}guez, Szep, Shao, Satam,
  Pacheco, and Hariri]{gao2022selecting}
Xin Gao, Sundaresh Ram, Rohit~C Philip, Jeffrey~J Rodr{\'\i}guez, Jeno Szep,
  Sicong Shao, Pratik Satam, Jes{\'u}s Pacheco, and Salim Hariri.
\newblock Selecting post-processing schemes for accurate detection of small
  objects in low-resolution wide-area aerial imagery.
\newblock \emph{Remote Sensing}, 14\penalty0 (2):\penalty0 255, 2022.

\bibitem[Howard et~al.(2019)Howard, Sandler, Chu, Chen, Chen, Tan, Wang, Zhu,
  Pang, Vasudevan, et~al.]{howard2019searching}
Andrew Howard, Mark Sandler, Grace Chu, Liang-Chieh Chen, Bo~Chen, Mingxing
  Tan, Weijun Wang, Yukun Zhu, Ruoming Pang, Vijay Vasudevan, et~al.
\newblock Searching for mobilenetv3.
\newblock In \emph{Proceedings of the IEEE/CVF international conference on
  computer vision}, pages 1314--1324, 2019.

\bibitem[Howard et~al.(2017)Howard, Zhu, Chen, Kalenichenko, Wang, Weyand,
  Andreetto, and Adam]{howard2017mobilenets}
Andrew~G Howard, Menglong Zhu, Bo~Chen, Dmitry Kalenichenko, Weijun Wang,
  Tobias Weyand, Marco Andreetto, and Hartwig Adam.
\newblock Mobilenets: Efficient convolutional neural networks for mobile vision
  applications.
\newblock \emph{arXiv preprint arXiv:1704.04861}, 2017.

\bibitem[Kapach et~al.(2012)Kapach, Barnea, Mairon, Edan, and
  Ben-Shahar]{kapach2012computer}
Keren Kapach, Ehud Barnea, Rotem Mairon, Yael Edan, and Ohad Ben-Shahar.
\newblock Computer vision for fruit harvesting robots--state of the art and
  challenges ahead.
\newblock \emph{International Journal of Computational Vision and Robotics},
  3\penalty0 (1/2):\penalty0 4--34, 2012.

\bibitem[Koch et~al.(2005)Koch, Dittmann, Lipa, Menzel, Nekovar, and van
  Vliet]{koch2005cost}
E~Koch, E~Dittmann, W~Lipa, A~Menzel, J~Nekovar, and AJH van Vliet.
\newblock Cost action 725: Establishing a european phenological data platform
  for climatological applications.
\newblock In \emph{17th International Congress of Biometeorology (ICB 20050),
  Offenbach am Main 2005}, pages 554--558, 2005.

\bibitem[Kollia et~al.(2021)Kollia, Stevenson, and Kollias]{kollia2021ai}
Ilianna Kollia, Jack Stevenson, and Stefanos Kollias.
\newblock Ai-enabled efficient and safe food supply chain.
\newblock \emph{Electronics}, 10\penalty0 (11):\penalty0 1223, 2021.

\bibitem[Lu and Young(2020)]{lu2020survey}
Yuzhen Lu and Sierra Young.
\newblock A survey of public datasets for computer vision tasks in precision
  agriculture.
\newblock \emph{Computers and Electronics in Agriculture}, 178:\penalty0
  105760, 2020.

\bibitem[Ma and He(2013)]{ma2013imbalanced}
Yunqian Ma and Haibo He.
\newblock Imbalanced learning: foundations, algorithms, and applications.
\newblock 2013.

\bibitem[Matvienko et~al.(2020)Matvienko, Gasanov, Petrovskaia, Jana,
  Pukalchik, and Oseledets]{matvienko2020bayesian}
Ivan Matvienko, Mikhail Gasanov, Anna Petrovskaia, Raghavendra~Belur Jana,
  Maria Pukalchik, and Ivan Oseledets.
\newblock Bayesian aggregation improves traditional single image crop
  classification approaches.
\newblock \emph{arXiv preprint arXiv:2004.03468}, 2020.

\bibitem[Mohanty et~al.(2016)Mohanty, Hughes, and
  Salath{\'e}]{mohanty2016using}
Sharada~P Mohanty, David~P Hughes, and Marcel Salath{\'e}.
\newblock Using deep learning for image-based plant disease detection.
\newblock \emph{Frontiers in plant science}, 7:\penalty0 1419, 2016.

\bibitem[Opitz and Burst(2019)]{opitz2019macro}
Juri Opitz and Sebastian Burst.
\newblock Macro f1 and macro f1.
\newblock \emph{arXiv preprint arXiv:1911.03347}, 2019.

\bibitem[Owusu et~al.(2021)Owusu, Kuffer, Belgiu, Grippa, Lennert, Georganos,
  and Vanhuysse]{owusu2021towards}
Maxwell Owusu, Monika Kuffer, Mariana Belgiu, Tais Grippa, Moritz Lennert,
  Stefanos Georganos, and Sabine Vanhuysse.
\newblock Towards user-driven earth observation-based slum mapping.
\newblock \emph{Computers, environment and urban systems}, 89:\penalty0 101681,
  2021.

\bibitem[Salvi et~al.(2021)Salvi, Acharya, Molinari, and
  Meiburger]{salvi2021impact}
Massimo Salvi, U~Rajendra Acharya, Filippo Molinari, and Kristen~M Meiburger.
\newblock The impact of pre-and post-image processing techniques on deep
  learning frameworks: A comprehensive review for digital pathology image
  analysis.
\newblock \emph{Computers in Biology and Medicine}, 128:\penalty0 104129, 2021.

\bibitem[Sandler et~al.(2018)Sandler, Howard, Zhu, Zhmoginov, and
  Chen]{sandler2018mobilenetv2}
Mark Sandler, Andrew Howard, Menglong Zhu, Andrey Zhmoginov, and Liang-Chieh
  Chen.
\newblock Mobilenetv2: Inverted residuals and linear bottlenecks.
\newblock In \emph{Proceedings of the IEEE conference on computer vision and
  pattern recognition}, pages 4510--4520, 2018.

\bibitem[Santoso et~al.(2021)Santoso, Purnomo, Sulianto, and
  Choirun]{santoso2021machine}
I~Santoso, M~Purnomo, AA~Sulianto, and A~Choirun.
\newblock Machine learning application for sustainable agri-food supply chain
  performance: a review.
\newblock In \emph{IOP Conference Series: Earth and Environmental Science},
  volume 924, page 012059. IOP Publishing, 2021.

\bibitem[Schiller et~al.(2021)Schiller, Schmidtlein, Boonman,
  Moreno-Mart{\'\i}nez, and Kattenborn]{schiller2021deep}
Christopher Schiller, Sebastian Schmidtlein, Coline Boonman, Alvaro
  Moreno-Mart{\'\i}nez, and Teja Kattenborn.
\newblock Deep learning and citizen science enable automated plant trait
  predictions from photographs.
\newblock \emph{Scientific Reports}, 11\penalty0 (1):\penalty0 1--12, 2021.

\bibitem[Sima et~al.(2020)Sima, Loudjani, and Devos]{sima2020use}
A~Sima, P~Loudjani, and W~Devos.
\newblock Use of geotagged photographs in the frame of common agriculture
  policy checks.
\newblock 2020.

\bibitem[Soekhoe et~al.(2016)Soekhoe, Putten, and Plaat]{soekhoe2016impact}
Deepak Soekhoe, Peter van~der Putten, and Aske Plaat.
\newblock On the impact of data set size in transfer learning using deep neural
  networks.
\newblock In \emph{International symposium on intelligent data analysis}, pages
  50--60. Springer, 2016.

\bibitem[Soille et~al.(2018)Soille, Burger, De~Marchi, Kempeneers, Rodriguez,
  Syrris, and Vasilev]{soille2018versatile}
Pierre Soille, A~Burger, D~De~Marchi, Pieter Kempeneers, D~Rodriguez, Vassilis
  Syrris, and V~Vasilev.
\newblock A versatile data-intensive computing platform for information
  retrieval from big geospatial data.
\newblock \emph{Future Generation Computer Systems}, 81:\penalty0 30--40, 2018.

\bibitem[Sudars et~al.(2020)Sudars, Jasko, Namatevs, Ozola, and
  Badaukis]{sudars2020dataset}
Kaspars Sudars, Janis Jasko, Ivars Namatevs, Liva Ozola, and Niks Badaukis.
\newblock Dataset of annotated food crops and weed images for robotic computer
  vision control.
\newblock \emph{Data in brief}, 31:\penalty0 105833, 2020.

\bibitem[Tian et~al.(2020)Tian, Wang, Liu, Qiao, and Li]{tian2020computer}
Hongkun Tian, Tianhai Wang, Yadong Liu, Xi~Qiao, and Yanzhou Li.
\newblock Computer vision technology in agricultural automation—a review.
\newblock \emph{Information Processing in Agriculture}, 7\penalty0
  (1):\penalty0 1--19, 2020.

\bibitem[Tseng et~al.(2021)Tseng, Zvonkov, Nakalembe, and
  Kerner]{tseng2021cropharvest}
Gabriel Tseng, Ivan Zvonkov, Catherine~Lilian Nakalembe, and Hannah Kerner.
\newblock Cropharvest: A global dataset for crop-type classification.
\newblock In \emph{Thirty-fifth Conference on Neural Information Processing
  Systems Datasets and Benchmarks Track (Round 2)}, 2021.

\bibitem[USDA()]{usgscc}
USDA.
\newblock United states department of agriculture foreign agriculture service.
\newblock URL
  \url{https://ipad.fas.usda.gov/rssiws/al/crop_calendar/europe.aspx}.

\bibitem[van~der Velde et~al.(2019)van~der Velde, van Diepen, and
  Baruth]{vanderVelde2019}
M.~van~der Velde, C.A. van Diepen, and B.~Baruth.
\newblock The european crop monitoring and yield forecasting system:
  Celebrating 25 years of {JRC} {MARS} bulletins.
\newblock \emph{Agricultural Systems}, 168:\penalty0 56--57, January 2019.
\newblock \doi{10.1016/j.agsy.2018.10.003}.
\newblock URL \url{https://doi.org/10.1016/j.agsy.2018.10.003}.

\bibitem[Waldner et~al.(2017)Waldner, Hansen, Potapov, L{\"o}w, Newby,
  Ferreira, and Defourny]{waldner2017national}
Fran{\c{c}}ois Waldner, Matthew~C Hansen, Peter~V Potapov, Fabian L{\"o}w,
  Terence Newby, Stefanus Ferreira, and Pierre Defourny.
\newblock National-scale cropland mapping based on spectral-temporal features
  and outdated land cover information.
\newblock \emph{PloS one}, 12\penalty0 (8):\penalty0 e0181911, 2017.

\bibitem[Wu et~al.(2021)Wu, Chen, Zhao, Kang, and Ding]{wu2021review}
Zhangnan Wu, Yajun Chen, Bo~Zhao, Xiaobing Kang, and Yuanyuan Ding.
\newblock Review of weed detection methods based on computer vision.
\newblock \emph{Sensors}, 21\penalty0 (11):\penalty0 3647, 2021.

\bibitem[Zheng et~al.(2019)Zheng, Kong, Jin, Wang, Su, and
  Zuo]{zheng2019cropdeep}
Yang-Yang Zheng, Jian-Lei Kong, Xue-Bo Jin, Xiao-Yi Wang, Ting-Li Su, and Min
  Zuo.
\newblock Cropdeep: The crop vision dataset for deep-learning-based
  classification and detection in precision agriculture.
\newblock \emph{Sensors}, 19\penalty0 (5):\penalty0 1058, 2019.

\bibitem[Zhu et~al.(2021)Zhu, Spachos, Pensini, and Plataniotis]{zhu2021deep}
Lili Zhu, Petros Spachos, Erica Pensini, and Konstantinos~N Plataniotis.
\newblock Deep learning and machine vision for food processing: A survey.
\newblock \emph{Current Research in Food Science}, 4:\penalty0 233--249, 2021.

\end{thebibliography}

\newpage

\appendix
\newpage

\section{Appendix A}
%% put this in apendix

\begin{sidewaysfigure}

% \begin{figure*}[h]
    \centering\includegraphics[width=\linewidth]{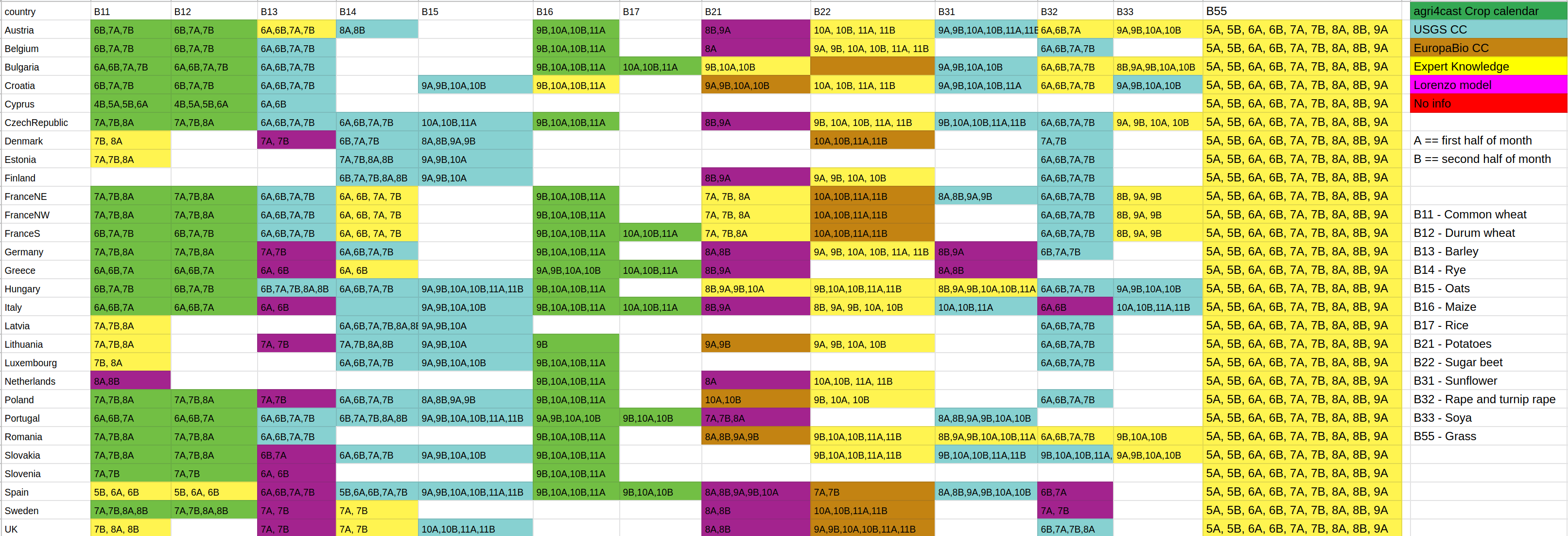}
\caption{Extracted harvest conditions of each crop for each country in the EU after CC harmonization and expert knowledge gap filling.}
\label{fig:finalTableHarvestStages}
% \end{figure*}

\end{sidewaysfigure}

% % latex table generated in R 4.2.1 by xtable 1.8-4 package
% % Wed Nov  9 23:25:56 2022
% % latex table generated in R 4.2.1 by xtable 1.8-4 package
% % Fri Nov 11 16:14:04 2022
% \begin{table}[ht]
% \centering
% \begin{tabular}{rlrrrrrr}
%   \hline
% ImageSets & MaxX & MinX & MaxY & MinY & RatioX & RatioY \\ 
%   \hline
% MMEC & 2976 & 480 & 5312 &  64 & 6.20 & 83.00 \\ 
% Train & 2580 & 480 & 5312 &  64 & 5.38 & 83.00 \\ 
% Test85 & 2580 & 768 & 4608 & 640 & 3.36 & 7.20 \\ 
% TestAll & 2976 & 480 & 5312 & 640 & 6.20 & 8.30 \\ 
%   \hline
% \end{tabular}
% \end{table}

% latex table generated in R 4.2.1 by xtable 1.8-4 package
% Mon Nov 14 17:23:45 2022
\begin{table}[ht]
\centering
\begin{tabular}{p{2.5cm} p{4cm} p{1cm}}
  \hline
 Range of pixels & WxH included & \% of images \\ 
  \hline
Less than 1 million & 640x480, 1024x768, 800x600  & 1.39 \\ 
1-2 million & 1600x1200, 1280x960, 1632x1224, 1200x1600, 1200x900, 1605x1204, 1728x1152, 1288x966, 1600x1198, 1612x1212, 1700x1130, 1600x963, 1261x817, 1600x900, 1397x1048, 1593x1200, 1319x989, 1280x1024, 1552x1164  & 64.73 \\ 
2-3 million & 1664x1248, 2048x1360, 1920x1080, 1824x1216, 1733x1300, 1792x1312, 1936x1288, 1656x1242, 1984x1488, 2048x1104, 2000x1333, 1824x1368, 1936x1296, 1936x1452, 1920x1440, 1662x1246, 2080x1368, 1360x2048, 2048x1376, 1800x1350, 1632x1232  & 6.68 \\ 
3-4 million & 2048x1536, 2304x1728, 2272x1704, 2288x1712, 1536x2048, 2352x1568, 2592x1458, 2200x1650, 2042x1532, 2133x1600, 2240x1680, 2080x1544  & 22.32 \\ 
4-5 million & 2560x1712, 2560x1920, 2400x1800, 2344x1758, 2464x1632, 1932x2580, 2576x1932  & 2.33 \\ 
5-6 million & 2592x1944, 2816x2112,  & 3.46 \\ 
6-7 million & 2848x2136, 3072x2048, 3456x1946, 2896x2172,  & 0.95 \\ 
7-8 million & 3072x2304, 3264x2448, 3584x2016  & 3.08 \\ 
 Over 8 million & 3968x2976, 3488x2616, 4320x2432, 3664x2748, 4672x3504, 3840x2880  & 0.81 \\ 
   \hline
\end{tabular}
    \caption{Breakdown of the kinds of image sizes present in the operational inference set (8642 images)}
    \label{tab:resolultions}
\end{table}

\begin{figure*}[!h]
    \centering\includegraphics[width=0.8\linewidth]{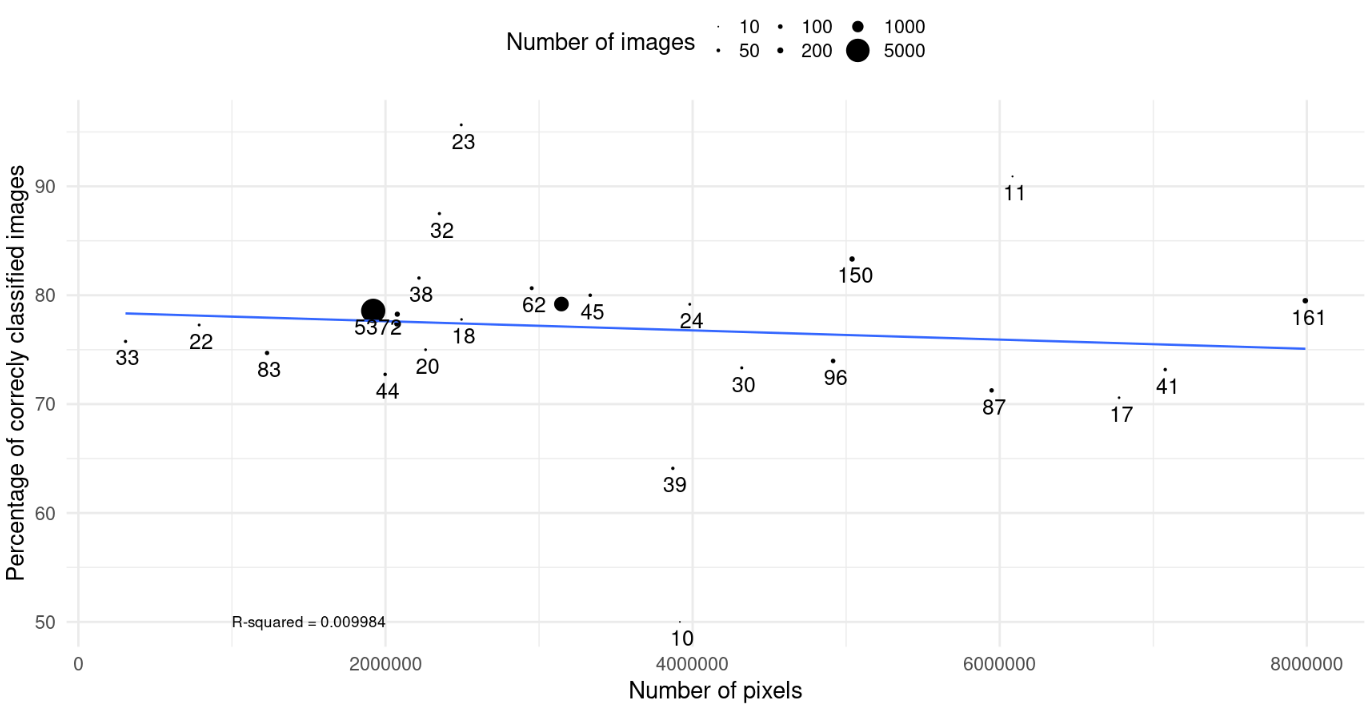}
\caption{Scatter-plot of the effects of image resolution, represented by the product of the image dimensions in numbers of pixels, on the validity of the classification, represented by the proportion of correctly classified examples in each resolution bin. The correlation between the two is given by the R-squared value at the bottom of the plot.}
\label{fig:lucasv_wxh_propr_point}
\end{figure*}






% \end{thebibliography}

\end{document}